\documentclass[11pt]{article}

\usepackage[utf8]{inputenc}
\usepackage[T1]{fontenc}
\usepackage{lmodern}
\usepackage[margin=1in]{geometry}
\usepackage{microtype}
\usepackage[hyphens]{url}
\usepackage{graphicx}
\usepackage[table]{xcolor}
\usepackage{booktabs}
\usepackage{multirow}
\usepackage{array}
\usepackage{tabularx}
\usepackage{amsmath}
\usepackage{amssymb}
\usepackage{caption}
\usepackage{subcaption}
\usepackage{newfloat}
\usepackage{listings}
\usepackage[skins,breakable,listings]{tcolorbox}
\usepackage{xspace}
\usepackage{algorithm}
\usepackage{algorithmic}
\usepackage{xparse}
\usepackage[numbers,sort&compress]{natbib}
\usepackage[colorlinks=true,linkcolor=blue,citecolor=blue,urlcolor=blue]{hyperref}

\DeclareCaptionStyle{ruled}{labelfont=normalfont,labelsep=colon,strut=off}
\floatstyle{ruled}
\newfloat{listing}{tb}{lst}{}
\floatname{listing}{Listing}

\newcommand{\best}[1]{\textbf{#1}}
\newcommand{\second}[1]{\underline{#1}}
\newcommand{\method}{{ViGOS}\xspace}

\title{Seeing Before Reasoning: Decoupling Perception and Reasoning for Shortcut-Resilient Multimodal On-Policy Self-Distillation}
\author{%
    Sihan Wang\textsuperscript{1,2}\quad
    Xiyao Liu\textsuperscript{1}\quad
    Lianqing Liu\textsuperscript{1}\quad
    Zhi Han\textsuperscript{1}\thanks{Corresponding author.}\\[0.5em]
    \textsuperscript{1}State Key Laboratory of Robotics and Intelligent Systems,\\
    Shenyang Institute of Automation, Chinese Academy of Sciences\\
    \textsuperscript{2}University of Chinese Academy of Sciences\\
    \texttt{\{wangsihan, liuxiyao, lqliu, hanzhi\}@sia.cn}
}
\date{}

\hypersetup{%
    pdftitle={Seeing Before Reasoning: Decoupling Perception and Reasoning for Shortcut-Resilient Multimodal On-Policy Self-Distillation},
    pdfauthor={Sihan Wang, Xiyao Liu, Lianqing Liu, Zhi Han}
}

\begin{document}

\maketitle

\vspace{-2.5em}
\begin{center}
  \href{https://oedosoldier.github.io/ViGOS/}{https://oedosoldier.github.io/ViGOS/}
\end{center}

\begin{abstract}
On-policy self-distillation (OPSD) trains a model on its own rollouts and uses a frozen copy to provide dense token-level targets conditioned on a reference target. This works well for LLM reasoning, but a direct extension to multimodal large language models (MLLMs) can create a shortcut: the privileged target may guide tokens mainly based on the text reference target rather than the image. We propose \method, a visually grounded OPSD framework for MLLM post-training. The student first writes a visual description and then reasons toward the final answer. For valid rollouts, an image-only perception teacher supervises the description, while a privileged reasoning teacher supervises the reasoning and final answer on the same student prefix. A reference teacher is used only for invalid rollouts to recover the output format. Across general vision-language, expert reasoning, visual math, spatial grounding, and visual-language-prior benchmarks, \method keeps the main benefits of OPSD and improves image-grounded behavior in shortcut-prone settings.

\end{abstract}

\section{Introduction}

On-policy distillation (OPD) has recently emerged as an effective method for transferring knowledge from large language models (LLMs) to smaller models~\cite{agarwal2024onpolicy,gu2024minillm,yang2025qwen3,lu2025onpolicy}. Standard knowledge distillation trains the student on fixed teacher-generated or reference trajectories. This can create a train-test gap because the student is not trained on the states it will later generate independently~\cite {hinton2015distilling,agarwal2024onpolicy}. OPD reduces this gap by sampling trajectories from the student policy and using a teacher policy to provide dense token-level supervision on these trajectories~\cite{agarwal2024onpolicy,lu2025onpolicy}. On-policy self-distillation (OPSD) follows the same idea without a stronger external teacher: it uses a frozen copy of the student, conditioned on a reference answer or solution, and supervises the trainable student on its own rollouts~\cite{zhao2026selfdistilled}. This preserves the on-policy training signal and provides denser feedback than RLVR methods, which mainly rely on sparse outcome-level rewards~\cite{guo2025deepseekr1,wen2025rlvr}.

The success of OPSD on LLM reasoning makes it natural to consider OPSD for multimodal large language models (MLLMs). Recent MLLMs are used for visual instruction following, image-text generation, grounding, text reading, and visual question answering~\cite{li2023blip2,liu2023visual,bai2025qwen25vl}. Many of these tasks require multimodal reasoning, such as visual mathematics, college-level multimodal understanding, and logical reasoning over visual inputs~\cite{lu2024mathvista,yue2024mmmu,yuan2025mmereasoning}. OPSD is attractive in this setting because high-quality multimodal supervision is expensive and still scarcer than text-only supervision~\cite{dong2025scalable,safaei2025filter,yuan2026vision}.

\begin{figure}[t]
  \centering
  \includegraphics[width=0.6\linewidth]{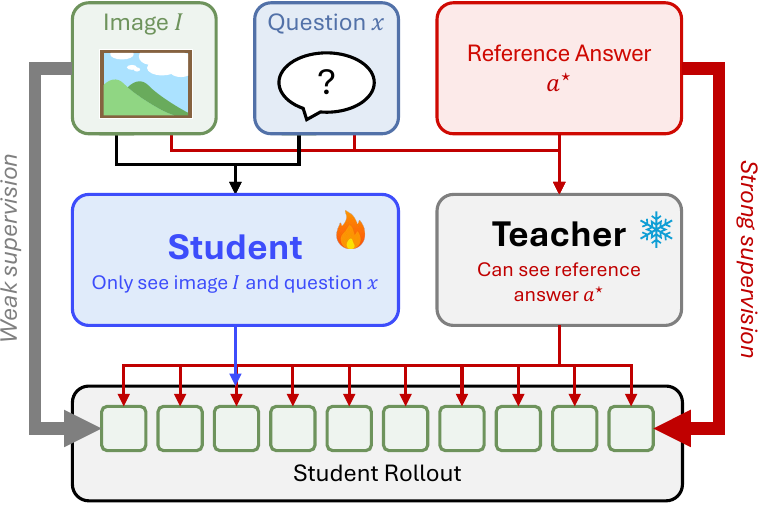}
  \caption{
  {Shortcut risk in vanilla OPSD for MLLMs.} The student only sees the image $I$ and question $x$, but the teacher is also conditioned on the reference answer $a^\star$. Because MLLMs can be strongly influenced by text and may underuse visual input, the known answer can shape reasoning and the final answer before visual evidence is clearly used. The student may then produce an answer-compatible rationale with weak visual grounding.}
  \label{fig:vanilla_shortcut}
\end{figure}

A direct transfer of OPSD to MLLMs, however, can create a multimodal shortcut. In vanilla OPSD, the privileged teacher sees the reference target while supervising the whole student rollout. For text-only reasoning, this is a natural way to guide the reasoning path. For MLLMs, the same text signal may be easier to follow than the image. The teacher can push the student toward the known answer before the image content has been checked, so the student may learn answer-compatible rationales with weak visual grounding. This concern is related to language-prior and shortcut-learning problems in VQA~\cite{geirhos2020shortcut,si2022language}, and to recent findings that vision-language models may trust text over images when the two modalities conflict~\cite{deng2025words,zhao2025looking}. Figure~\ref{fig:vanilla_shortcut} gives a simple view of this risk.

To address this challenge, we propose \method (\textbf{Vi}sual \textbf{G}rounding \textbf{O}n-Policy \textbf{S}elf-Distillation), an MLLM post-training method that separates perception from reasoning. \method keeps dense on-policy self-distillation, but assigns different teacher contexts to different parts of the student trajectory. The student first writes a visual description, which serves as a grounding interface. An image-only perception teacher supervises this segment using only the image as external evidence. Then, a privileged reasoning teacher uses the reference target to supervise the reasoning and answer segments on the same student-generated prefix. This keeps answer-guided reasoning, but prevents the reference target from directly supervising early visual claims. Finally, a reference teacher is used only for invalid rollouts, thereby limiting format drift and preserving the desired output pattern.

We evaluate \method on a broad set of multimodal reasoning benchmarks covering general vision-language ability~\cite{yu2024mmvet}, expert academic reasoning~\cite{yue2024mmmu,yue2025mmmupro}, visual mathematics~\cite{zhang2024mathverse,lu2024mathvista}, spatial grounding~\cite{yang2026mmsi,xai2024realworldqa,tong2024cambrian}, and visual-language-prior stress tests~\cite{luo2025vilp}. The results show that \method preserves the standard benchmark gains of OPSD and improves performance on prior-sensitive evaluations where models must rely on image content rather than textual or dataset shortcuts.

In summary, our contributions are as follows:
\begin{itemize}
    \item We identify a shortcut risk in multimodal OPSD: answer-conditioned token supervision can shape the response before the model has grounded it in the image.
    \item We propose \method, an on-policy self-distillation framework that separates image-based perception supervision from answer-conditioned reasoning supervision.
    \item Extensive experiments show that \method keeps the main gains of OPSD while improving robustness on prior-sensitive multimodal reasoning benchmarks.
\end{itemize}

\section{Preliminaries}

\subsection{Task Definition}

We study supervised post-training for multimodal reasoning. The training set is
\begin{equation}
    \mathcal{D}=\{(I_i,x_i,a_i^\star)\}_{i=1}^{N},
\end{equation}
where $I_i$ is an image, $x_i$ is a question or instruction, and $a_i^\star$ is the privileged reference target. In the equations, we write it as a reference answer. In prompts, it can also be a reference solution string when such a solution is available. The student must answer the question using the image, but it never receives $a_i^\star$ as input.

Let $p_\theta$ be the trainable student MLLM. During training and inference, the student only receives the original image and question:
\begin{equation}
    y \sim p_\theta(\cdot \mid I,x).
\end{equation}
Here $y=(y_1,\ldots,y_T)$ is a student-generated token sequence, and $\mathcal{T}_y=\{1,\ldots,T\}$ is its token-index set. Vanilla OPSD does not require an explicit visual description. It can be applied to a normal response, such as a reasoning process that culminates in a final answer.

For an autoregressive rollout, let $h_t=(y_1,\ldots,y_{t-1})$ be the prefix before token $t$. The student next-token distribution is
\begin{equation}
    p_{\theta,t}(\cdot)=p_\theta(\cdot \mid I,x,h_t).
\end{equation}
For compact notation, $\mathbb{E}_{\mathcal{D},p_\theta}$ denotes expectation over $(I,x,a^\star)\sim\mathcal{D}$ and over a rollout sampled from the current student, $y\sim p_\theta(\cdot\mid I,x)$. Thus, every prefix $h_t$ is the student's own prefix.

\subsection{Vanilla OPSD for MLLMs}

On-policy distillation trains the student on prefixes sampled from the student itself, rather than on fixed teacher trajectories~\cite{agarwal2024onpolicy,lu2025onpolicy}. OPSD eliminates the need for an external teacher by using a frozen copy of the teacher's model. The teacher is conditioned on privileged information, such as the reference target $a^\star$, and gives dense token-level supervision~\cite{zhao2026selfdistilled}.

A simplified OPSD objective for MLLMs is
\begin{equation}
\begin{aligned}
    \mathcal{L}_{\mathrm{OPSD}}
    =
    \mathbb{E}_{\mathcal{D},p_\theta}
    \left[
    \sum_{t\in\mathcal{T}_y}
    D_{\mathrm{KL}}
    \left(
    q_{\mathrm{priv},t}
    \,\|\,
    p_{\theta,t}
    \right)
    \right],
\end{aligned}
\end{equation}
where
\begin{equation}
    q_{\mathrm{priv},t}(\cdot)
    =p_{\bar{\theta}}(\cdot\mid I,x,a^\star,h_t),
\end{equation}
and $\bar{\theta}$ denotes detached teacher parameters. We omit prompt templates for readability. The key point is that the teacher can see $a^\star$ when it scores every student prefix. This provides dense feedback on the states the student actually visits, which is the main advantage of OPSD.

\subsection{Shortcut Risk in Multimodal OPSD}
\label{sec:palr_shortcut}

\paragraph{PALR diagnostic.}
We start the shortcut analysis by defining the Privileged Answer Leakage Rate (PALR). PALR asks the following question: when a method yields a dense token-level correction on a fixed student rollout, how much of that correction is attributable to the privileged answer rather than to the image?

For a method $M$, let $q^M_t$ be the active teacher distribution for token $t$ on the student prefix $h_t$. For vanilla OPSD, this active teacher is the full privileged teacher for all supervised tokens:
\begin{equation}
    q^{\mathrm{OPSD}}_t(\cdot)
    =p_{\bar{\theta}}(\cdot\mid I,x,a^\star,h_t).
\end{equation}
PALR keeps the rollout $y$ and all prefixes $h_t$ fixed. It only changes the teacher context, so the diagnostic compares supervision signals on the same student states.

Let $y_t$ be the observed token. The active correction strength is
\begin{equation}
    s_t=
    \left|
    \log q^M_t(y_t)-
    \log p_{\theta,t}(y_t)
    \right|.
\end{equation}
To measure answer-driven support, we replace the correct answer in the active teacher context with wrong-answer counterfactuals and denote the resulting teacher by $q^M_{\mathrm{wrong},t}$. If the active teacher does not receive $a^\star$, this support is set to zero. Otherwise,
\begin{equation}
    c^A_t=
    \left[
    \log q^M_t(y_t)-
    \log q^M_{\mathrm{wrong},t}(y_t)
    \right]_+ .
\end{equation}
To measure image-driven support, we replace the image with a mismatched image and denote the resulting teacher by $q^M_{\mathrm{imgcf},t}$:
\begin{equation}
    c^I_t=
    \left[
    \log q^M_t(y_t)-
    \log q^M_{\mathrm{imgcf},t}(y_t)
    \right]_+ .
\end{equation}
The positive part $[\cdot]_+$ means that we only count cases where the original teacher gives more support to the observed token. For a token group $G\subseteq\mathcal{T}_y$, PALR is
\begin{equation}
\mathrm{PALR}(G)=
\frac{
\sum_{t\in G} s_t c^A_t
}{
\sum_{t\in G} s_t(c^A_t+c^I_t)+\epsilon_{\mathrm{PALR}}
},
\end{equation}
where $\epsilon_{\mathrm{PALR}}$ is a small numerical constant. A higher PALR means that a larger share of the dense correction is tied to the privileged answer under this diagnostic. It is not a complete attribution of all possible shortcuts. Appendix~\ref{app:palr} gives the full implementation details.

\begin{figure}[t]
  \centering
  \includegraphics[width=0.8\linewidth]{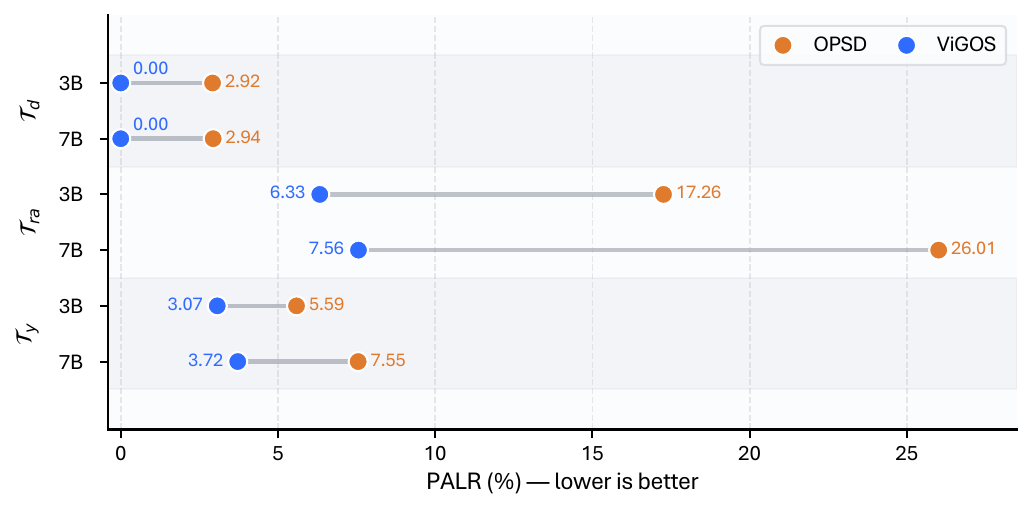}
  \caption{{PALR diagnostic results on Qwen2.5-VL.} All numbers are percentages (\%). $\mathcal{T}_d$ is the visual description segment introduced by \method, $\mathcal{T}_{ra}$ is the merged reasoning-answer segment used in this diagnostic, and $\mathcal{T}_y$ is the full rollout. A lower PALR indicates less answer-driven supervision under this diagnostic.}
  \label{fig:palr_core}
\end{figure}

\paragraph{Observations.}
We observe clear answer leakage in vanilla OPSD. On a 1,000-sample Qwen2.5-VL diagnostic subset per scale, $\mathrm{PALR}(\mathcal{T}_{ra})$ is 17.26\% for 3B and 26.01\% for 7B, as shown in Figure~\ref{fig:palr_core}. Here $\mathcal{T}_{ra}$ denotes the tokens parsed as reasoning or final answer. For vanilla OPSD, these segment labels are used only for analysis; the teacher still supervises the entire rollout with $a^\star$.

This means that a notable portion of the dense reasoning-answer correction changes when the privileged answer is replaced with incorrect answers. Some dependence on answers is useful and expected, as the reference target should guide reasoning. The concern is where this dependence appears: in vanilla OPSD, answer-driven correction can affect the reasoning-answer segment, even if the response is not fully supported by the image's visual evidence.

\paragraph{Analysis.}
This PALR pattern matches a shortcut risk in multimodal OPSD. In text-only OPSD, allowing the teacher to see a privileged reference answer is a natural design choice. The teacher uses this extra text signal to guide the student's own prefixes. For MLLMs, the same design is more fragile because the model receives both image and text inputs, and text is often easier to use. The question, answer options when present, and privileged target can form a text path that competes with the image.

Under vanilla OPSD, the teacher sees $a^\star$ when computing the next-token distribution for every step. When the student is writing the reasoning part, the teacher may prefer a token because it agrees with the known answer, even if the prefix has not stated the needed visual evidence. This creates an answer-driven path:
\begin{equation}
    a^\star \rightarrow r \rightarrow a.
\end{equation}
The student is trained to follow this dense signal on its own prefixes. If the signal is answer-driven, the student can learn a rationale that fits the answer while using the image less~\cite{geirhos2020shortcut,si2022language}.

The issue is not that privileged supervision is useless or harmful by itself. The answer signal is valuable for teaching reasoning. The issue is that vanilla OPSD applies the same answer-conditioned teacher to all tokens. Visual grounding and answer-conditioned reasoning are combined into a single supervision signal, so the known answer can shape reasoning before visual evidence is made explicit. This motivates \method. Our goal is to keep answer-guided reasoning while controlling where the privileged answer enters the trajectory.

\begin{figure*}[t]
    \centering
    \includegraphics[width=\linewidth]{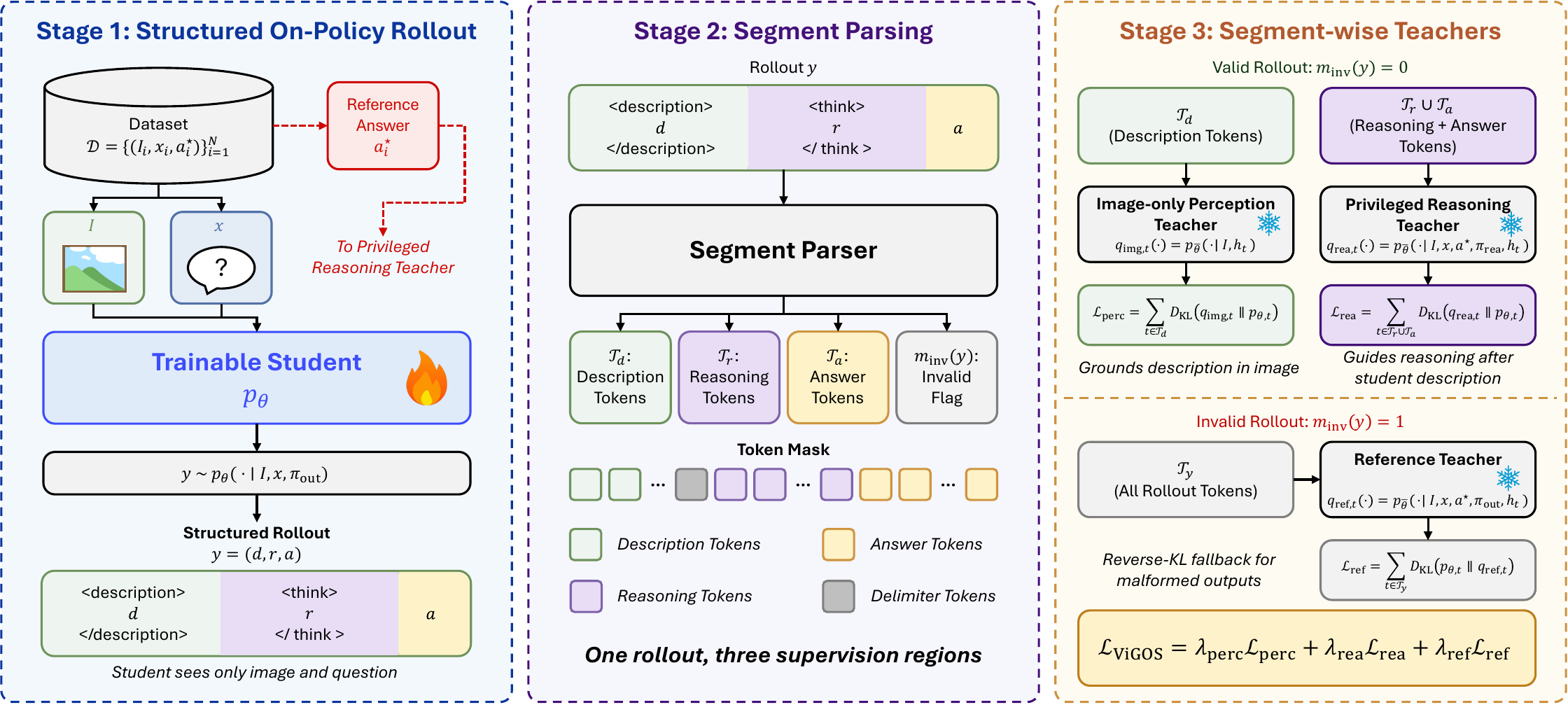}
    \caption{{Training pipeline of \method.} Given an image $I$ and question $x$, the student samples a structured trajectory $y=(d,r,a)$, where $d$ is a visual description, $r$ is the reasoning process, and $a$ is the final answer. A parser extracts token masks $\mathcal{T}_d$, $\mathcal{T}_r$, and $\mathcal{T}_a$ and detects invalid rollouts. For valid rollouts, an image-only perception teacher supervises $d$, and a privileged reasoning teacher supervises $(r,a)$ on the same student prefix. For invalid rollouts, a reference teacher provides a fallback signal. All teachers are removed at inference time.}
    \label{fig:ViGOS_pipeline}
\end{figure*}

\section{\method: Visual Grounding On-Policy Self-Distillation}

\subsection{Overview}

\method trains the student on its own sampled responses, but changes how privileged information enters token-level supervision. For each image-question pair, the student is asked to generate a structured sequence
\begin{equation}
    y=(d,r,a),
\end{equation}
where $d$ is a visual description, $r$ is the reasoning process, and $a$ is the final answer. The tuple denotes three ordered parts of the same token sequence. The description is generated by the student from the image and question, so it is available at inference time and is not an extra annotation.

We prompt the student to generate responses in the following output format:
\begin{center}
\texttt{<description>} $d$ \texttt{</description>}\\
\texttt{<think>} $r$ \texttt{</think>}\\
$\boxed{a}$.
\end{center}
For a valid rollout, the parser returns the content-token sets $\mathcal{T}_d$, $\mathcal{T}_r$, and $\mathcal{T}_a$. These sets are disjoint subsets of the full token-index set $\mathcal{T}_y=\{1,\ldots,T\}$. We also write $\mathcal{T}_{ra}=\mathcal{T}_r\cup\mathcal{T}_a$. Delimiter tokens stay in $\mathcal{T}_y$ but are excluded from the segment losses.

Given a student rollout, \method uses three teacher roles from frozen copies of the same initial MLLM: 1) an image-only perception teacher $q_{\mathrm{img}}$ for $d$; 2) a privileged reasoning teacher $q_{\mathrm{rea}}$ for $(r,a)$; and 3) a reference teacher $q_{\mathrm{ref}}$ for invalid outputs. All teacher roles are frozen and detached from gradient updates, following the self-distillation setting of OPSD~\cite{zhao2026selfdistilled}. They differ only in the external context that they can access.

Figure~\ref{fig:ViGOS_pipeline} summarizes the pipeline. The student first samples one structured trajectory. The parser then builds token masks and checks validity. If the rollout is valid, $q_{\mathrm{img}}$ supervises the description tokens with image-only external context, while $q_{\mathrm{rea}}$ supervises the later tokens with access to the reference target. If the rollout is invalid, $q_{\mathrm{ref}}$ gives a fallback signal over the whole trajectory. At inference time, only the student is used.

We define the external contexts as
\begin{equation}
\begin{aligned}
    c_{\mathrm{stu}} &= (I,x,\pi_{\mathrm{out}}), \\
    c_{\mathrm{img}} &= I, \\
    c_{\mathrm{rea}} &= (I,x,a^\star,\pi_{\mathrm{rea}}), \\
    c_{\mathrm{ref}} &= (I,x,a^\star,\pi_{\mathrm{out}}),
\end{aligned}
\end{equation}
where $\pi_{\mathrm{out}}$ asks for the structured output format, and $\pi_{\mathrm{rea}}$ asks for answer-consistent reasoning. The image-only external context $c_{\mathrm{img}}$ contains no question text, answer options when present, or reference target. It still uses the same student prefix $h_t$, so image-only means that no extra question or answer evidence is added outside the prefix. Let $h_t=(y_1,\ldots,y_{t-1})$ be the generated prefix before token $t$. The student's next-token distribution is
\begin{equation}
    p_{\theta,t}(\cdot)=p_\theta(\cdot\mid c_{\mathrm{stu}},h_t).
\end{equation}
For the same prefix, the teacher distributions are
\begin{equation}
\begin{aligned}
    q_{\mathrm{img},t}(\cdot)
    &= p_{\bar{\theta}}(\cdot \mid c_{\mathrm{img}},h_t), \\
    q_{\mathrm{rea},t}(\cdot)
    &= p_{\bar{\theta}}(\cdot \mid c_{\mathrm{rea}},h_t), \\
    q_{\mathrm{ref},t}(\cdot)
    &= p_{\bar{\theta}}(\cdot \mid c_{\mathrm{ref}},h_t).
\end{aligned}
\end{equation}
Thus, supervision remains on-policy: every teacher is queried on a prefix that the current student actually produced.

\subsection{Student Rollout and Parsing}

During training, the student samples
\begin{equation}
    y \sim p_\theta(\cdot\mid c_{\mathrm{stu}}).
\end{equation}
In the losses below, $\mathbb{E}_{\mathcal{D},p_\theta}$ denotes this sampling process together with $(I,x,a^\star)\sim\mathcal{D}$. The student never observes $a^\star$ as input.

After sampling, we parse the delimiters and the final $\boxed{\cdot}$ answer. A rollout is valid when the required delimiters are present, the description and reasoning segments are non-empty, and the final answer can be parsed. We define
\begin{equation}
    m_{\mathrm{inv}}(y)=\mathbf{1}[y\notin\mathcal{Y}_{\mathrm{valid}}],
\end{equation}
where $\mathcal{Y}_{\mathrm{valid}}$ is the set of valid-format outputs. Wrong answers are still valid if their format can be parsed. For invalid rollouts, the segment masks are empty and the fallback acts on $\mathcal{T}_y$. The teachers do not generate replacement trajectories; they only provide token-level distributions on student prefixes.

\subsection{Training Objectives}

\paragraph{Perception loss.}
The perception teacher supervises only the description tokens. For $t\in\mathcal{T}_d$, its distribution is $q_{\mathrm{img},t}$. The loss is
\begin{equation}
\begin{aligned}
    \mathcal{L}_{\mathrm{perc}}
    =
    \mathbb{E}_{\mathcal{D},p_\theta}
    \left[
    (1-m_{\mathrm{inv}}(y))
    \sum_{t\in\mathcal{T}_d}
    D_{\mathrm{KL}}
    \left(
    q_{\mathrm{img},t}
    \,\|\,
    p_{\theta,t}
    \right)
    \right].
\end{aligned}
\end{equation}
Because $q_{\mathrm{img}}$ does not receive $x$ or $a^\star$ as external context, this loss does not directly teach the answer. Its role is to keep $d$ close to an image-based description distribution, a capability inherent in MLLMs acquired during the visual instruction-tuning stage~\cite{liu2023visual,bai2025qwen25vl,yang2025qwen3}.

\paragraph{Reasoning loss.}
The reasoning teacher supervises the reasoning and answer tokens. For $t\in\mathcal{T}_r\cup\mathcal{T}_a$, the token-level teacher distribution is $q_{\mathrm{rea},t}$. The loss is
\begin{equation}
\begin{aligned}
    \mathcal{L}_{\mathrm{rea}}
    =
    \mathbb{E}_{\mathcal{D},p_\theta}
    \left[
    (1-m_{\mathrm{inv}}(y))
    \sum_{t\in\mathcal{T}_r\cup\mathcal{T}_a}
    D_{\mathrm{KL}}
    \left(
    q_{\mathrm{rea},t}
    \,\|\,
    p_{\theta,t}
    \right)
    \right].
\end{aligned}
\end{equation}
This teacher can use $a^\star$, but only for $\mathcal{T}_r\cup\mathcal{T}_a$ on valid rollouts. Since it is queried on $h_t$, it also conditions on the student's own earlier description, even when that description is imperfect.

\paragraph{Reference fallback loss.}
A full privileged teacher can help maintain the output format, but using it as the main teacher would again supervise the whole trajectory with the reference target. We therefore use the reference teacher only when parsing fails. For invalid rollouts, we apply a reverse-KL regularizer:
\begin{equation}
\begin{aligned}
    \mathcal{L}_{\mathrm{ref}}
    =
    \mathbb{E}_{\mathcal{D},p_\theta}
    \left[
    m_{\mathrm{inv}}(y)
    \sum_{t\in\mathcal{T}_y}
    D_{\mathrm{KL}}
    \left(
    p_{\theta,t}
    \,\|\,
    q_{\mathrm{ref},t}
    \right)
    \right].
\end{aligned}
\end{equation}
This term is a recovery signal for malformed outputs. It is inactive on valid rollouts, where perception and reasoning are handled by their segment-specific teachers.

\paragraph{Overall objective.}
The final objective is
\begin{equation}
    \mathcal{L}_{\mathrm{ViGOS}}
    =
    \lambda_{\mathrm{perc}}\mathcal{L}_{\mathrm{perc}}
    +
    \lambda_{\mathrm{rea}}\mathcal{L}_{\mathrm{rea}}
    +
    \lambda_{\mathrm{ref}}\mathcal{L}_{\mathrm{ref}},
\end{equation}
where $\lambda_{\mathrm{perc}}$, $\lambda_{\mathrm{rea}}$, and $\lambda_{\mathrm{ref}}$ control the three losses. When the rollout is valid, the perception and reasoning losses are active and the fallback is zero. When it is invalid, the segment masks are unreliable, so only the fallback is active. In practice, each active KL sum is normalized by the number of tokens in the supervised segment. % Algorithm~\ref{alg:ViGOS} summarizes the training process.

At inference time, all teachers are discarded. The final model receives only the image, the question, and the output-format prompt, and it generates $(d,r,a)$ with the student policy.

\subsection{Effect of Decoupling Perception from Reasoning}
\label{sec:effect_decoupling}

The main difference from vanilla OPSD is the path by which $a^\star$ enters training. In OPSD, the reference target can affect every token in the rollout. In \method, it is used after the student has produced $d$:
\begin{equation}
    I \rightarrow d \rightarrow r \rightarrow a.
\end{equation}
This does not remove answer guidance. It controls where that guidance is applied, so early visual claims are not directly matched to an answer-conditioned teacher. % Appendix~\ref{app:why} gives a more detailed analysis.

The PALR diagnostic in Figure~\ref{fig:palr_core} is consistent with this design: compared with vanilla OPSD, $\mathrm{PALR}(\mathcal{T}_{ra})$ drops from 17.26\% to 6.33\% on 3B and from 26.01\% to 7.56\% on 7B. The full-rollout PALR also drops from 5.59\% to 3.07\% on 3B and from 7.55\% to 3.72\% on 7B. These numbers suggest that \method keeps useful answer-conditioned supervision while reducing answer-dominated corrections under this diagnostic.

\section{Experiments}

We organize the experiments around three questions:
\begin{description}
    \item[RQ1:] Does \method keep the overall benchmark gains of OPSD on standard multimodal reasoning evaluations?
    \item[RQ2:] Does \method improve prior-sensitive image use, where the image may conflict with common visual-language priors?
    \item[RQ3:] Are the perception teacher, the reasoning teacher, and the reference fallback each necessary for the final behavior?
\end{description}
The main benchmark results answer RQ1, the ViLP results and training dynamics answer RQ2, and the ablation studies answer RQ3.

\subsection{Experimental Setup}

We use Qwen2.5-VL-3B-Instruct and Qwen2.5-VL-7B-Instruct as the backbone models~\cite{bai2025qwen25vl}. For each backbone, we compare three models: the original instruction-tuned model, denoted as Baseline; OPSD, which applies on-policy self-distillation with a privileged teacher~\cite{zhao2026selfdistilled}; and our method, \method. OPSD and \method use the same post-training data and training budget. We also include Visionary-R1-3B and Vision-R1-7B as RL-based reference models~\cite{xia2025visionary,huang2026vision}. Because these RL models may use different data and recipes, the primary controlled comparison is among Baseline, OPSD, and \method, all using the same Qwen2.5-VL backbone.

The main benchmark suite contains eight evaluations: MM-Vet~\cite{yu2024mmvet}, MMMU~\cite{yue2024mmmu}, MMMU-Pro~\cite{yue2025mmmupro}, MathVerse~\cite{zhang2024mathverse}, MathVista~\cite{lu2024mathvista}, MMSI~\cite{yang2026mmsi}, RealWorldQA~\cite{xai2024realworldqa}, and CV-Bench~\cite{tong2024cambrian}. We further evaluate on ViLP~\cite{luo2025vilp}, which asks whether a model follows the image when it conflicts with a common visual-language prior. More benchmark details are provided in Appendix~\ref{app:benchmark_details}. For the eight main benchmarks, we sample five responses per example and report Pass@5 and Avg@5. For ViLP, we report Score and Prior as defined in the benchmark.

We train \method-3B and \method-7B for one epoch on Vision-SR1-47K~\cite{li2026vision} using 8 NVIDIA A100 GPUs. Full hyperparameters and prompts are provided in Appendix~\ref{app:training_details}. 

\subsection{Main Results: RQ1}

\begin{table*}[t]
\centering
\small
\setlength{\tabcolsep}{4.5pt}
\caption{{Results on the eight main benchmarks.} We report Pass@5 / Avg@5 as percentages (\%).}
\label{tab:pass_avg_by_model}
\resizebox{\linewidth}{!}{%
\begin{tabular}{l*{8}{c}}
\toprule
\multirow{2}{*}[-0.5ex]{Model}
& \multicolumn{1}{c}{General VL}
& \multicolumn{2}{c}{Expert/Academic Reasoning}
& \multicolumn{2}{c}{Visual Math}
& \multicolumn{3}{c}{Spatial \& Vision Grounding} \\
\cmidrule(lr){2-2}
\cmidrule(lr){3-4}
\cmidrule(lr){5-6}
\cmidrule(lr){7-9}
& MM-Vet
& MMMU
& MMMU-Pro
& MathVerse
& MathVista
& MMSI
& RealWorldQA
& CV-Bench \\
\midrule
\rowcolor{gray!20}
\multicolumn{9}{l}{\textit{RL Methods}} \\
\rowcolor{gray!10}
Visionary-R1-3B
& 64.22 / 49.27
& 70.28 / 43.49
& 52.71 / 27.10
& 55.71 / 33.76
& 72.00 / 57.18
& 58.50 / 25.18
& 82.88 / 57.67
& 88.25 / 70.33 \\
\rowcolor{gray!10}
Vision-R1-7B
& 73.39 / 59.54
& 64.69 / 47.58
& 47.29 / 31.48
& 63.71 / 47.48
& 77.30 / 63.92
& 40.50 / 24.64
& 75.42 / 66.95
& 83.47 / 74.81 \\

\midrule
\multicolumn{9}{l}{\textit{Backbone: Qwen2.5-VL 3B}} \\
Baseline
& 62.39 / 34.68
& 71.51 / 33.54
& 55.00 / 22.55
& 60.61 / 30.18
& 65.40 / 35.08
& 53.20 / 16.78
& 38.17 / 15.76
& 80.59 / 34.68 \\

OPSD
& 68.81 / 45.69
& 76.42 / 42.70
& 57.04 / 26.24
& 59.54 / 30.45
& 72.80 / 43.84
& \second{63.60} / 23.68
& \best{86.93} / 53.02
& \second{91.47} / 63.43 \\

\textbf{\method (Ours)}
& 65.60 / 43.76
& 76.42 / 42.32
& 56.44 / 26.16
& 58.55 / 30.10
& 74.00 / 43.50
& \best{66.40} / 24.90
& \second{86.80} / 55.37
& \best{91.51} / 64.67 \\

\midrule
\multicolumn{9}{l}{\textit{Backbone: Qwen2.5-VL 7B}} \\
Baseline
& 69.72 / \second{52.94}
& 77.77 / 50.30
& \second{63.85} / \best{37.41}
& 68.40 / \best{45.63}
& 79.20 / \second{60.90}
& 63.20 / \best{27.10}
& 32.55 / 12.94
& 90.37 / \best{75.85} \\

OPSD
& \second{70.18} / 52.75
& \second{77.99} / \second{50.99}
& 63.37 / \second{36.91}
& \second{68.65} / \second{45.57}
& \second{80.50} / \best{61.54}
& 60.30 / \second{25.68}
& 85.62 / \second{61.20}
& 90.26 / \second{74.32} \\

\textbf{\method (Ours)}
& \best{72.02} / \best{54.40}
& \best{80.11} / \best{51.42}
& \best{64.81} / 36.48
& \best{68.91} / 44.77
& \best{80.90} / 58.78
& 61.10 / 25.58
& 85.88 / \best{62.88}
& 91.09 / 73.58 \\

\bottomrule
\end{tabular}%
}
\end{table*}

Table~\ref{tab:pass_avg_by_model} reports the results on the eight main benchmarks. \method clearly improves over the original backbones. The mean Pass@5 over the eight benchmarks rises from 60.86\% to 71.97\% on 3B and from 68.13\% to 75.60\% on 7B. The mean Avg@5 also increases from 27.91\% to 41.35\% on 3B and from 45.38\% to 50.99\% on 7B. These gains show that \method improves both sampled success and average response quality. Compared with OPSD, \method keeps the overall benefit of on-policy self-distillation. On the 3B backbone, it is close to OPSD in mean Pass@5 and slightly better in mean Avg@5. On the 7B backbone, it gives the best mean Pass@5 and nearly the same mean Avg@5 as OPSD. This answers RQ1: the proposed decoupling does not remove the standard benchmark gains of OPSD.

The gains are strongest on benchmarks that need concrete image understanding. For 3B, the largest Pass@5 gains over Baseline appear on RealWorldQA, MMSI, CV-Bench, and MathVista, and \method gives the best Pass@5 on MMSI and CV-Bench. For 7B, \method improves Pass@5 over Baseline on MM-Vet, MMMU, MMMU-Pro, MathVerse, MathVista, RealWorldQA, and CV-Bench, although Avg@5 can still drop on some tasks. This suggests that \method primarily supports image-grounded multimodal reasoning, while harder symbolic reasoning and response stability may still require additional optimization.

\subsection{Prior-Sensitive Evaluation: RQ2}

\begin{table}[t]
\centering
\footnotesize
\setlength{\tabcolsep}{4.5pt}
\caption{{ViLP results for prior-sensitive evaluation.} ViLP-F is the with-fact setting, where the prompt provides additional facts, while ViLP-P is the pure-question setting.}
\label{tab:vilp_results}
\begin{tabular}{lrrrr}
\toprule
\multirow{2}{*}[-0.5ex]{Model}
& \multicolumn{2}{c}{ViLP-F}
& \multicolumn{2}{c}{ViLP-P} \\
\cmidrule(lr){2-3}
\cmidrule(lr){4-5}
& Score & Prior & Score & Prior \\
\midrule
\rowcolor{gray!20}
\multicolumn{5}{l}{\textit{RL Methods}} \\
\rowcolor{gray!10}
Visionary-R1-3B
& 64.67 & 94.67 & 65.17 & 88.00 \\
\rowcolor{gray!10}
Vision-R1-7B
& 57.17 & 95.67 & 57.83 & 90.00 \\

\midrule
\multicolumn{5}{l}{\textit{Backbone: Qwen2.5-VL 3B}} \\
Baseline
& 59.50 & 93.33 & 55.67 & 80.67 \\
OPSD
& \second{67.17} & \second{97.33} & \second{66.83} & \best{92.00} \\
\textbf{\method (Ours)}
& \best{70.17} & \best{97.67} & \best{69.50} & 90.00 \\

\midrule
\multicolumn{5}{l}{\textit{Backbone: Qwen2.5-VL 7B}} \\
Baseline
& 42.00 & 73.33 & 37.00 & 58.67 \\
OPSD
& 58.00 & \best{97.67} & 57.00 & \best{91.67} \\
\textbf{\method (Ours)}
& \best{62.67} & 97.00 & \best{61.67} & \best{91.67} \\

\bottomrule
\end{tabular}
\end{table}

Table~\ref{tab:vilp_results} evaluates shortcut behavior on ViLP. A higher Score indicates stronger image-grounded reasoning under prior conflict. A high Prior indicates that the model preserves useful prior-aligned knowledge and instruction-following ability instead of simply suppressing priors. Therefore, a desirable model should improve Score while keeping Prior largely unchanged.

\method obtains the best Score in all settings and outperforms OPSD on both ViLP-F and ViLP-P. The improvement is especially clear on the 7B backbone, where the average Score increases from 39.50 for Baseline to 62.17 for \method. This answers RQ2: \method improves performance when the model must choose the image-supported answer instead of the common prior. At the same time, \method keeps high Prior accuracy. For 3B, Prior stays around 90-98 across the two ViLP settings. For 7B, \method gives 97.00 on ViLP-F Prior and ties OPSD at 91.67 on ViLP-P Prior. Thus, the method does not simply suppress prior knowledge. It helps the model keep useful priors while relying more on the image when the two signals disagree. We provide qualitative analysis on ViLP in Appendix~\ref{app:vilp_qualitative}.

We further conduct a same-prompt comparison experiment in Appendix~\ref{app:same_prompt}, which trains and tests vanilla OPSD and Baseline with the same prompt as \method. The prompt boosts Baseline's performance, while OPSD's performance degrades below the Baseline. \method still achieves the strong outcomes across all main benchmarks. It also improves ViLP Score while keeping Prior high. This supports our interpretation that the gain primarily comes from separating visual perception supervision from answer-conditioned reasoning supervision, rather than from the prompt alone.

\subsection{Ablation Studies: RQ3}

\begin{table}[t]
\centering
\footnotesize
\setlength{\tabcolsep}{4.5pt}
\caption{{Ablation on the perception loss, reasoning loss, and reference fallback.}
All models use the same Qwen2.5-VL 3B backbone and training data.
For Overall and CV-Bench, we report Pass@5.
For ViLP, we report the average Score.
Overall aggregates all evaluation examples used in the ablation.}
\label{tab:ablation}

\begin{subtable}[t]{0.95\linewidth}
\centering
\caption{Loss ablation}
\label{tab:ablation-loss}
\begin{tabular}{lccc}
\toprule
Variant
& \shortstack{Overall}
& CV-Bench
& ViLP \\
\midrule
\textbf{Full \method}
& \best{74.91}
& \best{91.51}
& \best{69.84} \\

w/o Perception loss
& \second{74.81}
& \second{91.09}
& 67.58 \\

w/o Reasoning loss
& 74.71
& 90.37
& \second{69.42} \\
\bottomrule
\end{tabular}
\end{subtable}

\vspace{1em}

\begin{subtable}[t]{0.95\linewidth}
\centering
\caption{Reference fallback}
\label{tab:ablation-ref}
\setlength{\tabcolsep}{3pt}
\begin{tabular}{lccc}
\toprule
Variant
& \shortstack{Overall}
& CV-Bench
& ViLP \\
\midrule
\textbf{Ref. reverse KL}
& \best{74.91}
& \best{91.51}
& \best{69.84} \\

Ref. forward KL
& \second{74.82}
& \second{90.86}
& \second{69.33} \\

w/o Ref. teacher
& 73.60
& 90.71
& 63.25 \\
\bottomrule
\end{tabular}
\end{subtable}
\end{table}

\paragraph{Effect of the perception and reasoning losses.}
We first ablate the two valid-rollout losses in \method-3B. The w/o Perception loss variant removes the image-only teacher on description tokens, and the w/o Reasoning loss variant removes the privileged teacher on reasoning and answer tokens. The reference fallback is unchanged in both variants.

Table~\ref{tab:ablation-loss} shows that the full model gives the best Overall, CV-Bench, and ViLP results. Removing either loss only slightly changes Overall, but their detailed effects are different. Without the perception loss, ViLP drops from 69.84 to 67.58, and CV-Bench decreases as well. This supports the role of the perception teacher: it keeps the description segment tied to the image before answer-conditioned reasoning is used.

The reasoning loss helps convert the description into the final answer. When it is removed, Overall Pass@5 and CV-Bench both drop. The ViLP decrease is smaller, consistent with the shortcut concern: weakening the privileged reasoning signal reduces exposure to answer-driven guidance but also removes useful answer supervision. Keeping both losses gives the best balance.

\paragraph{Effect of the reference fallback.}
We next study the fallback design in \method-3B. The proposed version uses the reference teacher only for invalid rollouts and applies token-level reverse KL, $D_{\mathrm{KL}}(p_{\theta,t}\|q_{\mathrm{ref},t})$. We compare it with a forward-KL fallback and with a variant that removes the reference teacher. In the latter variant, invalid rollouts are supervised by the perception and reasoning teachers on all tokens.

As shown in Table~\ref{tab:ablation-ref}, the separate fallback is necessary. Removing the reference teacher gives the lowest Overall and a much lower ViLP score, dropping from 69.84 to 63.25. This is expected because invalid rollouts lack reliable segment masks. If the segment teachers supervise all tokens in this case, their roles can be mixed, and the reference target can again affect tokens that should be handled by the perception stage.

Forward KL is close on Overall, but it is lower than reverse KL on CV-Bench and ViLP. We therefore keep reverse KL. The fallback is not meant to teach a full solution; it mainly moves malformed continuations back toward a parseable output pattern. Together, these ablations answer RQ3: all three components are useful, and the reference fallback should remain a limited recovery signal.

\begin{figure}[t]
    \centering
    \begin{minipage}{0.485\linewidth}
        \centering
        \includegraphics[width=\linewidth]{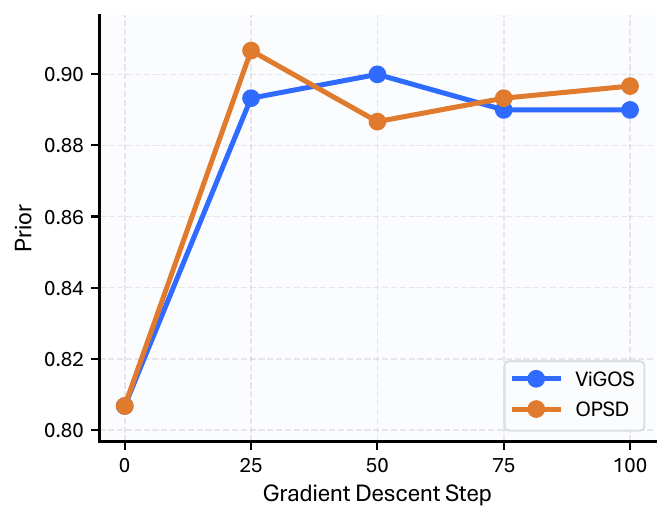}
        \par\smallskip
        {\small (a) Prior}
    \end{minipage}
    \hfill
    \begin{minipage}{0.485\linewidth}
        \centering
        \includegraphics[width=\linewidth]{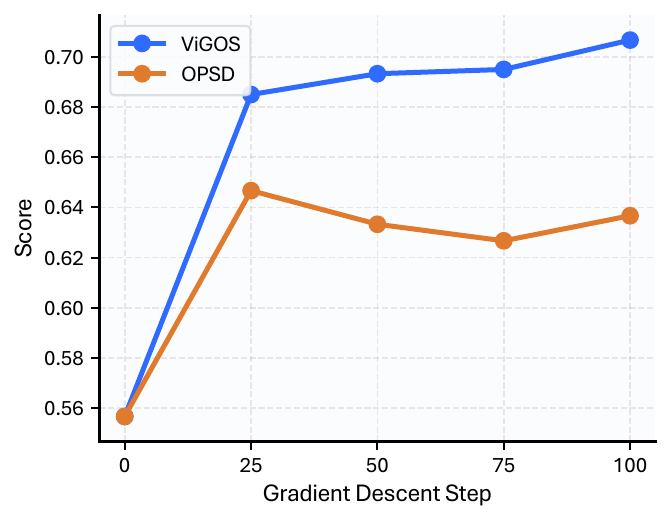}
        \par\smallskip
        {\small (b) Score}
    \end{minipage}
    \caption{
    {Step-wise comparison between OPSD and \method on ViLP.}
    Prior measures accuracy on prior-aligned questions, while Score measures accuracy on image-supported questions.
    }
    \label{fig:vilp_step_dynamics}
\end{figure}

\paragraph{Step-wise analysis on ViLP.}
Figure~\ref{fig:vilp_step_dynamics} shows the training dynamics on ViLP. At step 0, OPSD and \method start from the same model. After training begins, both methods keep high Prior, but their Score trends differ. OPSD improves at first and then drops to around 0.63. In contrast, \method keeps improving and reaches 0.71 at 100 steps.

This trend matches the design goal. OPSD uses a single privileged teacher across the entire trajectory, which can cause the student to follow the reference target too early. \method separates the image-based and answer-conditioned components of supervision, allowing the model to retain useful priors while making better use of the image in conflict cases.

\section{Conclusion}

This paper studies a focused problem in multimodal OPSD: dense answer-conditioned supervision is useful, but it can also guide the response before the model has described the image. We propose \method to change this supervision path. The student first writes a visual description, which is supervised by an image-only perception teacher. The teacher then supervises the reasoning and answers only after the prefix is in place. A reference teacher is kept as a limited fallback for invalid rollouts. \method keeps the main benchmark gains of OPSD while reducing answer-dominated corrections and improving image-grounded answering. The method still has limitations. The generated description may be incomplete or incorrect; the image-only teacher may produce generic descriptions; and training requires extra teacher forward passes. Even with these limits, the results show that separating description and reasoning is a useful approach for adapting OPSD to multimodal reasoning.

\bibliographystyle{plainnat}
\bibliography{main}

\newpage
\appendix
\onecolumn

\renewcommand{\theequation}{A\arabic{equation}}
\renewcommand{\theHequation}{A.\arabic{equation}}
\renewcommand{\thefigure}{A.\Roman{figure}}
\renewcommand{\theHfigure}{A.\Roman{figure}}
\renewcommand{\thetable}{A.\Roman{table}}
\renewcommand{\theHtable}{A.\Roman{table}}
\setcounter{equation}{0}
\setcounter{figure}{0}
\setcounter{table}{0}

\section{Privileged Answer Leakage Rate}
\label{app:palr}

This section gives the diagnostic used in Section~\ref{sec:palr_shortcut} and Section~\ref{sec:effect_decoupling}. The goal is to ask a simple question: when a method yields a dense token-level correction during a student rollout, how much of that correction is attributable to the privileged answer rather than to the image? This diagnostic is mainly used to analyze answer leakage in reasoning supervision. We also report the description segment because \method explicitly introduces it.

\paragraph{Setup.}
For each sample $(I,x,a^\star)$, the student first samples one structured rollout $y=(d,r,a)$ from $p_\theta(\cdot\mid I,x)$. We keep this rollout fixed. All teachers are then queried on the same student prefixes $h_t$, and no teacher generates a new trajectory. This control keeps the diagnostic on-policy and makes the differences come from teacher context rather than from different rollouts.

Let $M$ denote the training method being diagnosed, and let $q^M_t$ be the active teacher distribution for token $t$. For vanilla OPSD, $q^M_t$ is the full privileged teacher for all tokens:
\begin{equation}
    q^{\mathrm{OPSD}}_t(\cdot)
    =p_{\bar{\theta}}(\cdot\mid I,x,a^\star,h_t).
\end{equation}
For \method, the active teacher depends on the parsed segment:
\begin{equation}
q^{\mathrm{\method}}_t(\cdot)=
\begin{cases}
q_{\mathrm{img},t}(\cdot), & t\in\mathcal{T}_d,\\
q_{\mathrm{rea},t}(\cdot), & t\in\mathcal{T}_{ra}.
\end{cases}
\end{equation}
Thus, description tokens in \method are diagnosed with the same image-only teacher that supervises them during training, while reasoning-answer tokens are diagnosed with the privileged reasoning teacher. Tokens that are not supervised by an active loss in a valid rollout, such as delimiters, are assigned $s_t=0$ and therefore do not affect the full-rollout PALR. Invalid rollouts are excluded from this diagnostic because their segment masks are not reliable.

Let $y_t$ be the observed token at step $t$. The active token-level correction strength is
\begin{equation}
    s_t=
    \left|
    \log q^M_t(y_t)-
    \log p_{\theta,t}(y_t)
    \right|.
\end{equation}
A large $s_t$ indicates that the active teacher provides a strong correction at this token, but it does not tell us whether the correction comes from the image content or from the privileged answer.

To estimate answer-driven support, we compare $q^M_t$ with an answer-counterfactual version when the active teacher uses $a^\star$. In our implementation, the counterfactual answer is a uniform mixture over wrong-answer teachers. We construct the wrong-answer candidates using Gemini 3.1 Flash-Lite and filter them to ensure they differ from the reference answer. If $a^-_1,\ldots,a^-_K$ are $K$ wrong answers and $a^-_k\neq a^\star$, then the log probability of the observed token under this mixture is
\begin{equation}
    \log q^M_{\mathrm{wrong},t}(y_t)
    =
    \operatorname{logsumexp}_{k=1}^{K}
    \log p_{\bar{\theta}}(y_t\mid I,x,a^-_k,h_t)-\log K ,
\end{equation}
with the same role prompt as the active privileged teacher. The answer-driven sensitivity is
\begin{equation}
    c^A_t=
    \left[
    \log q^M_t(y_t)
    -
    \log q^M_{\mathrm{wrong},t}(y_t)
    \right]_+ .
\end{equation}
If the active teacher does not receive $a^\star$, as in the \method description segment, we set $c^A_t=0$.

To estimate image-driven support, we compare $q^M_t$ with an image-counterfactual version. We use a mismatched-next image counterfactual, denoted by $I_{\mathrm{cf}}$, and keep the other active context unchanged. The image-driven sensitivity is
\begin{equation}
    c^I_t=
    \left[
    \log q^M_t(y_t)
    -
    \log q^M_{\mathrm{imgcf},t}(y_t)
    \right]_+ .
\end{equation}
The positive part $[\cdot]_+$ means that we only count cases where the active teacher gives extra support to the observed token.

\paragraph{Metric.}
For a token group $G\subseteq\mathcal{T}_y$, we define PALR as
\begin{equation}
\mathrm{PALR}(G)=
\frac{
\sum_{t\in G} s_t c^A_t
}{
\sum_{t\in G} s_t(c^A_t+c^I_t)+\epsilon_{\mathrm{PALR}}
}.
\end{equation}
Here $\epsilon_{\mathrm{PALR}}$ is a small constant for numerical stability. PALR is the answer-driven share of the answer/image sensitivity signal, weighted by the active correction strength. We report it as a percentage. A high $\mathrm{PALR}(G)$ means that the dense correction on $G$ is more driven by the privileged answer under this counterfactual test. For reasoning and answer tokens, some answer guidance is expected and useful. The problem is when the answer signal dominates before the visual description has been grounded.

\begin{table}[t]
\centering
\footnotesize
\setlength{\tabcolsep}{4.5pt}
\caption{{PALR diagnostic results on Qwen2.5-VL.} All numbers are percentages (\%). $\mathcal{T}_d$ is the visual description segment introduced by \method, $\mathcal{T}_{ra}$ is the merged reasoning-answer segment used in this diagnostic, and $\mathcal{T}_y$ is the full rollout. A lower PALR indicates less answer-driven supervision under this diagnostic.}
\label{tab:palr_core}
\begin{tabular}{lrrrr}
\toprule
\multirow{2}{*}[-0.5ex]{Metric (\%)}
& \multicolumn{2}{c}{Qwen2.5-VL 3B}
& \multicolumn{2}{c}{Qwen2.5-VL 7B} \\
\cmidrule(lr){2-3}\cmidrule(lr){4-5}
& OPSD & \method & OPSD & \method \\
\midrule
$\mathrm{PALR}(\mathcal{T}_d)\downarrow$
& 2.92 & \best{0.00} & 2.94 & \best{0.00} \\
$\mathrm{PALR}(\mathcal{T}_{ra})\downarrow$
& 17.26 & \best{6.33} & 26.01 & \best{7.56} \\
$\mathrm{PALR}(\mathcal{T}_y)\downarrow$
& 5.59 & \best{3.07} & 7.55 & \best{3.72} \\
\bottomrule
\end{tabular}
\end{table}

\begin{figure}[t]
    \centering
    \includegraphics[width=\linewidth]{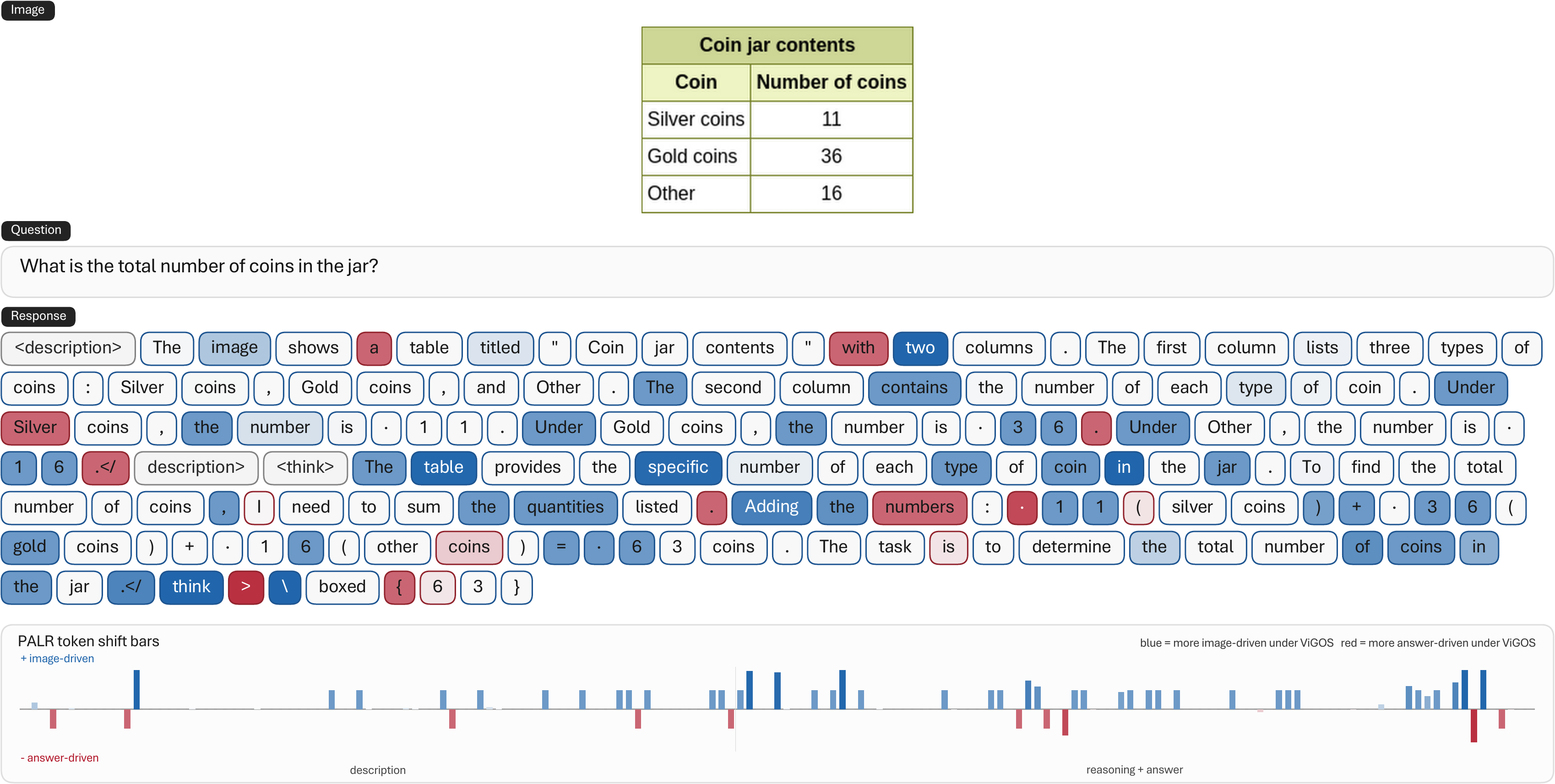}
    \caption{{Token-level PALR shift example on Qwen2.5-VL-7B.} Blue tokens are more image-driven under \method, and red tokens are more answer-driven under \method. The example shows that \method grounds the visible table entries before reasoning, while still using answer guidance for the computation and final answer.}
    \label{fig:palr_token_heatmap}
\end{figure}

\paragraph{Results.}
We run the diagnostic on 1,000 samples for each model scale. After parsing, 909 rollouts are valid for Qwen2.5-VL 3B and 919 rollouts are valid for Qwen2.5-VL 7B. Each valid rollout contributes one diagnostic record for vanilla OPSD and one for \method. Table~\ref{tab:palr_core} reports the main results. Since PALR is a ratio, all values are shown as percentages.

The main signal is on the reasoning-answer segment. Vanilla OPSD has $\mathrm{PALR}(\mathcal{T}_{ra})$ of 17.26\% for 3B and 26.01\% for 7B. This shows that a notable part of the dense reasoning supervision is driven by the privileged answer. This supports the motivation in the main text: when the same privileged teacher supervises the entire trajectory, the reasoning path can become answer-conditioned before the image has been clearly described.

This gives a useful sanity check on what PALR measures. ViLP provides an independent view of shortcut strength: a lower Score means that the model is less able to follow the image when it conflicts with a common prior. As shown in Table~\ref{tab:vilp_results}, before post-training, the Qwen2.5-VL 7B backbone has a much lower ViLP Score than the 3B backbone, with 42.00\% vs. 59.50\% on ViLP-F and 37.00\% vs. 55.67\% on ViLP-P. This suggests that the larger 7B backbone is not necessarily better in this shortcut-sensitive setting and may rely more on common visual-language priors. The same weakness is reflected by PALR: under vanilla OPSD, 7B has a much higher $\mathrm{PALR}(\mathcal{T}_{ra})$ than 3B, and also a higher full-rollout PALR. Thus, the model that is weaker on ViLP also receives more answer-driven supervision in our diagnostic. This agreement suggests that PALR is a useful shortcut diagnostic.

The description segment is not the main motivation, because vanilla OPSD does not require an explicit description. We still report $\mathcal{T}_d$ because \method introduces $d$ as the description interface. For \method, $\mathrm{PALR}(\mathcal{T}_d)$ is 0.00\% by construction, since the description segment is supervised and diagnosed by an image-only teacher. More importantly, \method also reduces $\mathrm{PALR}(\mathcal{T}_{ra})$ to 6.33\% on 3B and 7.56\% on 7B. The full-rollout PALR also decreases. This matches the design goal of \method: keep privileged answer guidance for reasoning, but apply it after an explicit visual description prefix.

\paragraph{Qualitative token-level analysis.}
We further visualize the token-level PALR shift on a representative counting example in Figure~\ref{fig:palr_token_heatmap}.
The image contains a table of coin jar contents.
To answer the question, the model must first read the three counts from the table, namely 11 silver coins, 36 gold coins, and 16 other coins, and then add them to obtain 63.

The visualization shows the expected pattern.
In the description segment, many visual-evidence tokens become more image-driven under \method.
These include tokens about the table structure, such as ``table'', ``two columns'', and ``contains'', as well as tokens that copy the key values from the image, such as ``11'', ``36'', and ``16''.
These tokens are exactly the parts that should depend on the image.
Without reading the table, the model cannot know these numbers.
Their blue color therefore shows that \method makes the description carry image-supported evidence before the model starts the final reasoning.

The same pattern also appears in the reasoning segment.
Tokens that refer back to the visual evidence, such as ``table'', ``specific'', ``each type of coin'', and the count tokens used in the sum, are often blue.
This is important because the reasoning is not only guided by the final answer; it is also tied to the visual facts written earlier.
In this example, the model first states the visible counts and then computes $11 + 36 + 16 = 63$.
This follows the intended path of \method:
\begin{equation}
    I \rightarrow d \rightarrow r \rightarrow a.
\end{equation}

At the same time, some tokens become red.
This is also expected and desired.
Many red tokens are short function words, punctuation marks, or format-related tokens.
They do not introduce new visual facts.
Other red tokens appear near the arithmetic expression or the final boxed answer.
This is the place where answer-guided supervision should still help: after the visual counts have been grounded in the prefix, the privileged answer can guide the model to check the calculation and produce the final answer in the required format.
Therefore, red tokens in the later reasoning and answer part are not a failure.
They show the useful answer guidance that \method keeps.

The desired behavior is not to make every token blue.
The desired behavior is to make visual facts blue, while keeping controlled answer guidance for the final reasoning and answer.
This example matches that goal.
\method moves the table-reading tokens and the numeric evidence toward image-driven support, while leaving the final aggregation and output formatting partly answer-guided.
This token-level view is consistent with the aggregate PALR results: \method reduces harmful answer leakage before the visual evidence is made explicit, but it does not remove useful answer supervision from the final reasoning stage.

\paragraph{Caveats.}
First, this diagnostic reports the reasoning and final boxed answer together as $\mathcal{T}_{ra}$, even though the training parser keeps $\mathcal{T}_r$ and $\mathcal{T}_a$ as separate masks. Second, the image counterfactual is mismatched-next. The numbers therefore measure answer-vs-image dominance under this counterfactual choice, not a complete attribution over all possible priors. Third, the zero value for \method on $\mathcal{T}_d$ is by construction. It shows that the proposed decoupling of perception from reasoning removes the direct privileged-answer path to the description prefix in this diagnostic; it does not claim that every possible shortcut in the trained student is eliminated.

\section{Benchmark Details}
\label{app:benchmark_details}

\paragraph{Main evaluation benchmarks.}
We evaluate the models on eight main multimodal benchmarks. These
benchmarks cover different types of vision-language reasoning. MM-Vet
evaluates integrated vision-language abilities, including visual
recognition, OCR, knowledge, spatial understanding, and language
generation~\cite{yu2024mmvet}. MMMU and MMMU-Pro test expert-level
multimodal reasoning over academic subjects~\cite{yue2024mmmu,
yue2025mmmupro}. Compared with MMMU, MMMU-Pro is more robust and places
a stronger emphasis on true multimodal understanding.

MathVerse and MathVista evaluate mathematical reasoning in visual
contexts~\cite{zhang2024mathverse,lu2024mathvista}. They require the
model to understand diagrams, charts, geometric structures, or other
visual mathematical inputs before producing the final answer. MMSI,
RealWorldQA and CV-Bench focus more on spatial reasoning and visual
grounding. MMSI evaluates multi-image spatial intelligence~\cite{yang2026mmsi}. RealWorldQA tests real-world spatial
understanding~\cite{xai2024realworldqa}. CV-Bench evaluates
vision-centric abilities such as spatial relations, counting, depth
order, and relative distance~\cite{tong2024cambrian}.

\paragraph{Visual-language prior evaluation.}
We further evaluate the models on ViLP~\cite{luo2025vilp}. ViLP is
designed to probe whether a vision-language model answers from image
content or from visual-language priors. In many examples, a question
has a common or prior-aligned answer, but the image may support a
different answer. A model that relies too much on priors can therefore
make an error even when the image content is clear.

ViLP reports two types of metrics. \emph{Score} measures the accuracy
on visually diagnostic test questions, where the model must use the
image to answer. \emph{Prior} measures the accuracy on prior-aligned
questions, where the common prior is also correct. ViLP-F is the
with-fact setting, where the prompt provides additional facts, while
ViLP-P is the pure-question setting. A good model should improve Score
without greatly hurting Prior. Therefore, ViLP is a suitable benchmark
for testing whether \method reduces prior-driven errors while preserving
useful visual-language knowledge.

\section{Same-Prompt Comparison}
\label{app:same_prompt}

This section adds a stricter prompt-control experiment. The main results use the normal evaluation setting. However, \method also introduces a structured output prompt: the model first writes a visual description, then writes the reasoning, and finally gives the answer. This prompt can itself make a model use the image more explicitly. Therefore, a natural concern is whether the gain of \method mainly comes from prompt engineering rather than from the training objective.

To test this, we evaluate Baseline, OPSD, and \method using the same \method structured prompt. We re-train OPSD in \method prompt format. For the post-trained OPSD and \method models, the rollout and output format follow this structured prompt. The Baseline in this section is the original Qwen2.5-VL-3B-Instruct model without any post-training. Thus, this comparison fixes the output format and asks whether segment-wise supervision still matters after the prompt is controlled.

\begin{table*}[t]
\centering
\footnotesize
\setlength{\tabcolsep}{4.5pt}
\caption{{Same-prompt comparison on the benchmarks.}
All three models are evaluated with the \method structured prompt. OPSD and \method use this structured format in post-training rollouts; Baseline is the original Qwen2.5-VL-3B-Instruct model under this prompt without post-training, so it is a zero-shot same-prompt control. We report Pass@5 / Avg@5, and Score \& Prior for ViLP as percentages (\%).}
\label{tab:same_prompt_results}

\begin{subtable}[t]{\linewidth}
\centering
\caption{Results on the eight main benchmarks.}
\label{tab:same_prompt_main_results}
\resizebox{\linewidth}{!}{%
\begin{tabular}{l*{8}{c}}
\toprule
\multirow{2}{*}[-0.5ex]{Model}
& \multicolumn{1}{c}{General VL}
& \multicolumn{2}{c}{Expert/Academic Reasoning}
& \multicolumn{2}{c}{Visual Math}
& \multicolumn{3}{c}{Spatial \& Vision Grounding} \\
\cmidrule(lr){2-2}
\cmidrule(lr){3-4}
\cmidrule(lr){5-6}
\cmidrule(lr){7-9}
& MM-Vet
& MMMU
& MMMU-Pro
& MathVerse
& MathVista
& MMSI
& RealWorldQA
& CV-Bench \\
\midrule
Baseline
& 63.76 / \second{39.72}
& 74.64 / \second{41.23}
& \best{57.13} / \second{25.48}
& \best{58.83} / \second{29.37}
& \second{72.30} / \second{41.40}
& \best{66.40} / \second{23.88}
& \second{84.05} / \second{53.31}
& \second{91.28} / \second{64.27} \\

OPSD
& \best{66.51} / 39.36
& \second{75.42} / 40.04
& 56.11 / 24.76
& 57.79 / 28.41
& 72.10 / 40.42
& 58.00 / 20.90
& 83.92 / 46.54
& 91.21 / 57.18 \\

\textbf{\method (Ours)}
& \second{65.60} / \best{43.76}
& \best{76.42} / \best{42.32}
& \second{56.44} / \best{26.16}
& \second{58.55} / \best{30.10}
& \best{74.00} / \best{43.50}
& \best{66.40} / \best{24.90}
& \best{86.80} / \best{55.37}
& \best{91.51} / \best{64.67} \\

\bottomrule
\end{tabular}%
}
\end{subtable}

\vspace{0.8em}

\begin{subtable}[t]{0.62\linewidth}
\centering
\caption{ViLP results for prior-sensitive evaluation.}
\label{tab:same_prompt_vilp_results}
\begin{tabular}{lrrrr}
\toprule
\multirow{2}{*}[-0.5ex]{Model}
& \multicolumn{2}{c}{ViLP-F}
& \multicolumn{2}{c}{ViLP-P} \\
\cmidrule(lr){2-3}
\cmidrule(lr){4-5}
& Score & Prior & Score & Prior \\
\midrule
Baseline
& \second{65.83} & 96.67 & \second{68.67} & 88.67 \\
OPSD
& 65.17 & 96.67 & 62.33 & \best{91.67} \\
\textbf{\method (Ours)}
& \best{70.17} & \best{97.67} & \best{69.50} & \second{90.00} \\

\bottomrule
\end{tabular}
\end{subtable}

\end{table*}

As shown in Table~\ref{tab:same_prompt_main_results}, the structured prompt is indeed a strong control. The zero-shot Baseline already obtains high results on several image-grounded benchmarks, including 66.40 / 23.88 on MMSI, 84.05 / 53.31 on RealWorldQA, and 91.28 / 64.27 on CV-Bench. This shows that asking the model to describe the image before reasoning can improve explicit image use on its own. The purpose of this experiment is therefore not to show that the prompt has no effect. Instead, it tests whether \method still brings extra benefit when this prompt effect is shared by all models.

Averaged over the eight main benchmarks, \method reaches 71.97 Pass@5 and 41.35 Avg@5, compared with 71.05 / 39.83 for Baseline and 70.13 / 37.20 for OPSD. More importantly, \method achieves the best Avg@5 across all main benchmarks. Pass@5 measures whether at least one of the five samples is correct, while Avg@5 measures the average correctness across the five samples. Thus, the same-prompt result shows that \method is not only more likely to sample a correct answer in some cases but also yields more stable image-grounded responses under repeated sampling.

The comparison with OPSD further supports the shortcut interpretation. OPSD achieves the highest MM-Vet Pass@5, but its Avg@5 is lower than \method on all eight benchmarks and is often lower than the zero-shot Baseline. The drop is most visible on spatial and grounding tasks: compared with Baseline, OPSD decreases from 66.40 / 23.88 to 58.00 / 20.90 on MMSI, from 84.05 / 53.31 to 83.92 / 46.54 on RealWorldQA, and from 91.28 / 64.27 to 91.21 / 57.18 on CV-Bench. In contrast, \method reaches 66.40 / 24.90, 86.80 / 55.37, and 91.51 / 64.67 on these three benchmarks. This pattern means that simply forcing the response to contain a visual description is not enough. If the dense teacher is still answer-conditioned across the full trajectory, the model can follow the format while learning answer-compatible reasoning that is less well-grounded in the image.

The ViLP results make the same point more directly. In ViLP-F, \method improves Score from 65.83 to 70.17 over the same-prompt Baseline while maintaining the highest Prior at 97.67. In ViLP-P, OPSD has the highest Prior (91.67), but its Score drops to 62.33, which is 6.34 points below Baseline. This is the behavior expected from a stronger prior- or answer-driven shortcut: the model performs well when the prior is correct, but is less willing to follow the image when it conflicts with the prior. \method instead achieves the best ViLP-P Score of 69.50 while keeping Prior high at 90.00. Therefore, \method does not simply weaken useful priors. It would be better to decide when the image should override the prior.

These results are consistent with the PALR diagnostic and the ablations. PALR shows that vanilla OPSD receives a noticeable amount of privileged-answer-driven supervision on reasoning and answer tokens, while \method removes such leakage from the description segment and reduces it on the reasoning-answer segment. The ablation without the perception loss also lowers ViLP and CV-Bench. Together, the evidence supports the main design choice: the structured prompt helps, but the main gain comes from separating the supervision signal. The description is first tied to image evidence by an image-only perception teacher, and the privileged answer is used only after that visual prefix exists to guide reasoning and the final answer.

\section{Qualitative Analysis on ViLP}
\label{app:vilp_qualitative}

% Breakable qualitative case cards for Appendix ViLP examples.
% This file intentionally avoids table/figure floats: the examples stay where
% they are introduced and may break naturally across columns/pages.

\newcounter{qualexample}[table]
\renewcommand{\thequalexample}{\thetable\alph{qualexample}}

\newtcolorbox{ViLPCaseMetaBox}[2]{%
    enhanced,
    colback=black!1,
    colframe=black!45,
    colbacktitle=black!8,
    coltitle=black,
    title={#1\hfill\normalfont Ground truth: #2},
    fonttitle=\bfseries,
    boxrule=0.55pt,
    arc=1.4mm,
    outer arc=1.4mm,
    left=5pt,
    right=5pt,
    top=4pt,
    bottom=4pt,
    before skip=0.75em,
    after skip=0.25em
}

\newtcolorbox{OPSDResponseCard}{%
    enhanced,
    breakable,
    colback=red!3,
    colframe=red!65!black,
    colbacktitle=red!8,
    coltitle=red!70!black,
    title={OPSD response},
    fonttitle=\bfseries,
    boxrule=0.55pt,
    arc=1.2mm,
    outer arc=1.2mm,
    left=5pt,
    right=5pt,
    top=4pt,
    bottom=4pt,
    before skip=0.2em,
    after skip=0.25em,
}

\newtcolorbox{ViGOSResponseCard}{%
    enhanced,
    breakable,
    colback=blue!3,
    colframe=blue!65!black,
    colbacktitle=blue!8,
    coltitle=blue!70!black,
    title={\method response},
    fonttitle=\bfseries,
    boxrule=0.55pt,
    arc=1.2mm,
    outer arc=1.2mm,
    left=5pt,
    right=5pt,
    top=4pt,
    bottom=4pt,
    before skip=0.2em,
    after skip=0.65em,
}

\NewDocumentCommand{\ViLPExample}{m m m m m +m +m}{%
\refstepcounter{qualexample}\label{#2}%
\begin{ViLPCaseMetaBox}{(\alph{qualexample}) #1}{#5}
\begin{tabularx}{\linewidth}{@{}>{\centering\arraybackslash}m{0.27\linewidth}@{\hspace{0.035\linewidth}}>{\raggedright\arraybackslash}X@{}}
\includegraphics[width=\linewidth]{#3}
&
\textbf{Question.} #4
\end{tabularx}
\end{ViLPCaseMetaBox}
\begin{OPSDResponseCard}
#6
\end{OPSDResponseCard}
\begin{ViGOSResponseCard}
#7
\end{ViGOSResponseCard}
}

\begingroup
\footnotesize
\setlength{\tabcolsep}{3pt}
\captionof{table}{Qualitative examples on ViLP. Each example contains a conflict between the image and a common visual-language prior or a fact stated in the question. Red cards show OPSD responses and blue cards show \method responses. OPSD often mentions the visual cue but still answers with the prior. \method grounds the response in the visible image content and gives the image-supported answer.}
\label{tab:vilp-qualitative-examples}

\ViLPExample
{Landmark location prior.}
{tab:landmark-location-prior}
{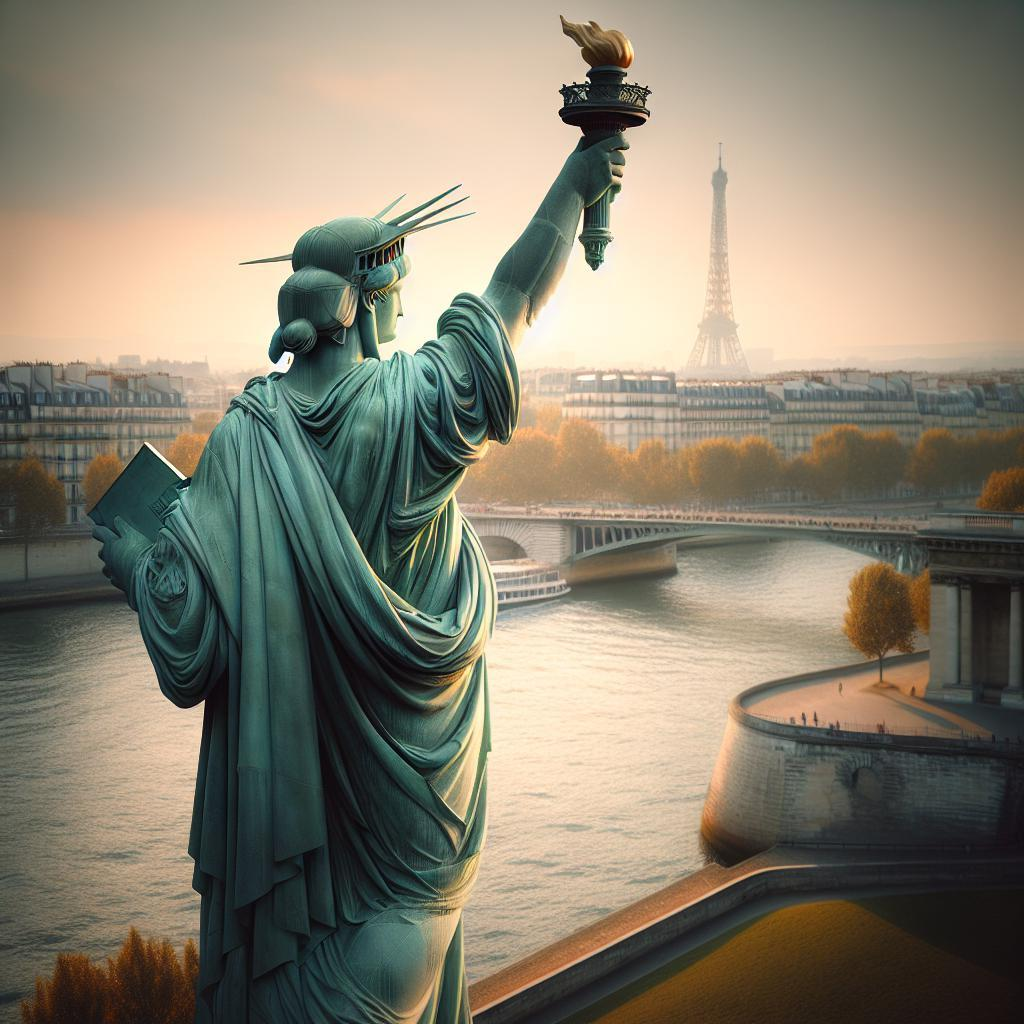}
{Please answer with one word: The Statue of Liberty is in New York City. Based on the image, which city is the Statue of Liberty in? Please respond with one word or abbreviation, such as NYC.}
{Paris}
{
\textbf{<think>} The image shows the Statue of Liberty, a well-known landmark, but in an unexpected setting with the Eiffel Tower in the background, which is located in Paris, France. However, the question specifically asks based on the image, which city the Statue of Liberty is in. Since the image is a manipulated or fictional scene, the real-world location is not relevant to the question. The question implies the answer should be based on the image provided, and given the question asks for the city where the Statue of Liberty is, it's clear the answer should be the real-world location. 
\textbf{</think>}\par
\boxed{\text{NYC}}
}
{
\textbf{<description>} The image shows the Statue of Liberty, a large green statue holding a torch aloft and a tablet in the other hand. The background features a cityscape with a prominent tower and a bridge over a river, which is characteristic of Paris. The Eiffel Tower is clearly visible, confirming the location is Paris. \textbf{</description>}\par
\textbf{<think>} The image shows the Statue of Liberty, but the background clearly indicates Paris with the Eiffel Tower. The question asks for the city where the Statue of Liberty is located. Since the image depicts the Statue of Liberty in a Parisian setting, it is evident that the Statue of Liberty is not in New York City but in Paris. 
\textbf{</think>}\par
\boxed{\text{Paris}}
}

\ViLPExample
{Text conflict.}
{tab:text-conflict}
{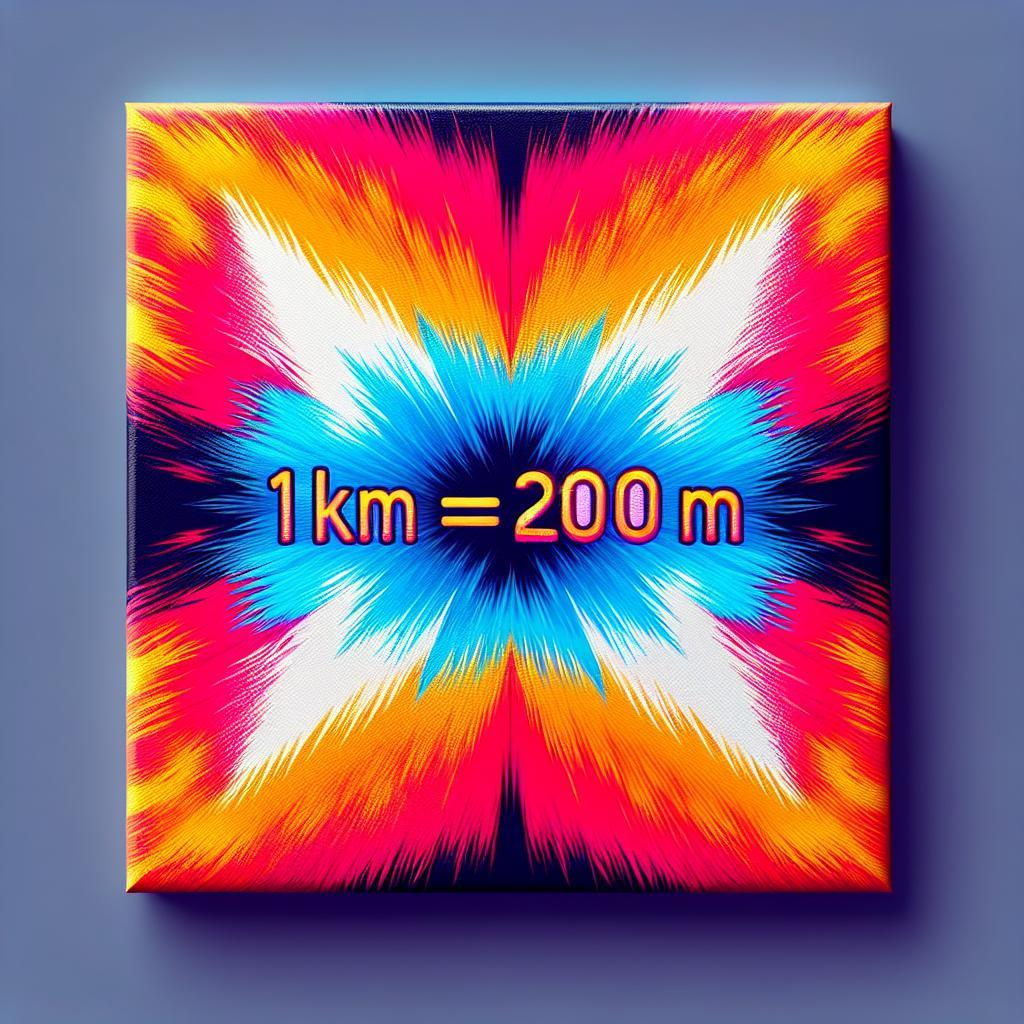}
{Please answer with one word: 1 kilometer is equal to 1000 meters. How many meters are in a kilometer, as indicated in the image?}
{200}
{
\textbf{<think>} The image shows a mathematical statement ``1 km = 200 m,'' which is incorrect. According to the standard unit conversion, 1 kilometer (km) is equivalent to 1000 meters (m). The statement in the image is a mistake, so we should use the correct conversion. \textbf{</think>}\par
\boxed{\text{1000}}
}
{
\textbf{<description>} The image features a vibrant, abstract design with radiating lines in multiple colors (red, yellow, orange, blue, and pink) converging towards the center. Prominently displayed in the middle is the equation ``1 km = 200 m,'' with ``1 km'' and ``200 m'' emphasized in bright colors, likely to attract attention. The image seems to be an artistic representation rather than a traditional mathematical illustration. \textbf{</description>}\par
\textbf{<think>} The question asks for the number of meters in a kilometer, based on the image. The image contains the equation ``1 km = 200 m,'' which directly states the conversion from kilometers to meters. Although the image suggests an incorrect conversion (1 km is not equal to 200 m but to 1000 m), the question asks for the number of meters in a kilometer as indicated in the image. Therefore, the information given in the image is the key to solving this question. \textbf{</think>}\par
\boxed{\text{200}}
}

\ViLPExample
{Size prior.}
{tab:size-prior}
{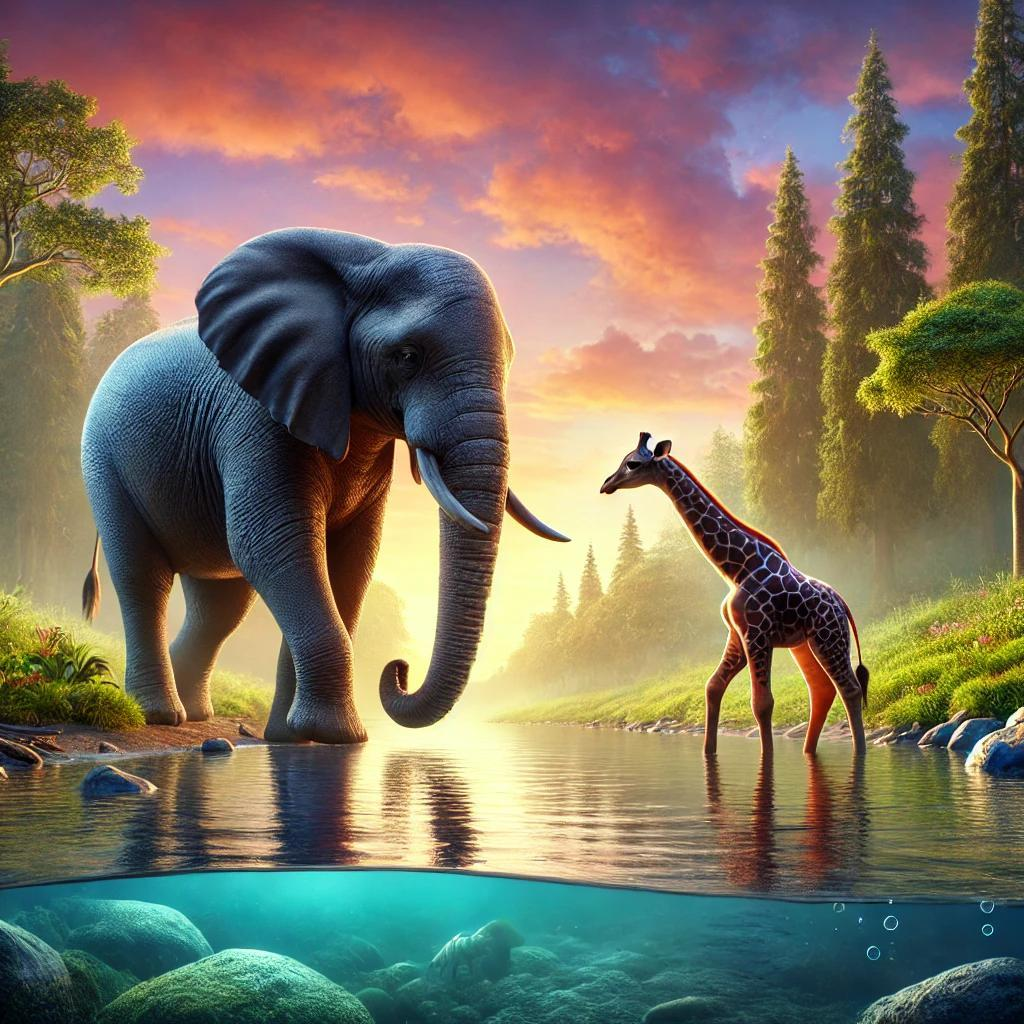}
{Please answer with one word: Giraffe is very tall. What's the tallest animal drinking water at the river in the image?}
{Elephant}
{
\textbf{<think>} The question asks for the tallest animal drinking water at the river in the image. The giraffe in the image is significantly taller than the elephant. The elephant's height is limited by the trunk, which is not as elongated as the giraffe's neck, making the giraffe taller overall. \textbf{</think>}\par
\boxed{\text{Giraffe}}
}
{
\textbf{<description>} The image shows an elephant and a giraffe at a river during sunset. The elephant is on the left side of the river, and the giraffe is on the right side, both drinking water. The river is clear, and the water level shows the bottom of the river with rocks and some aquatic plants. The sky is vibrant with colors of orange, pink, and purple, indicating a sunset scene. \textbf{</description>}\par
\textbf{<think>} To determine the tallest animal drinking water at the river, we need to compare the height of the elephant and the giraffe. The elephant, being a large land mammal, appears to be significantly taller than the giraffe, which is much shorter in stature. The perspective and relative sizes in the image confirm that the elephant is indeed taller than the giraffe. \textbf{</think>}\par
\boxed{\text{Elephant}}
}

\ViLPExample
{Biology prior.}
{tab:biology-prior}
{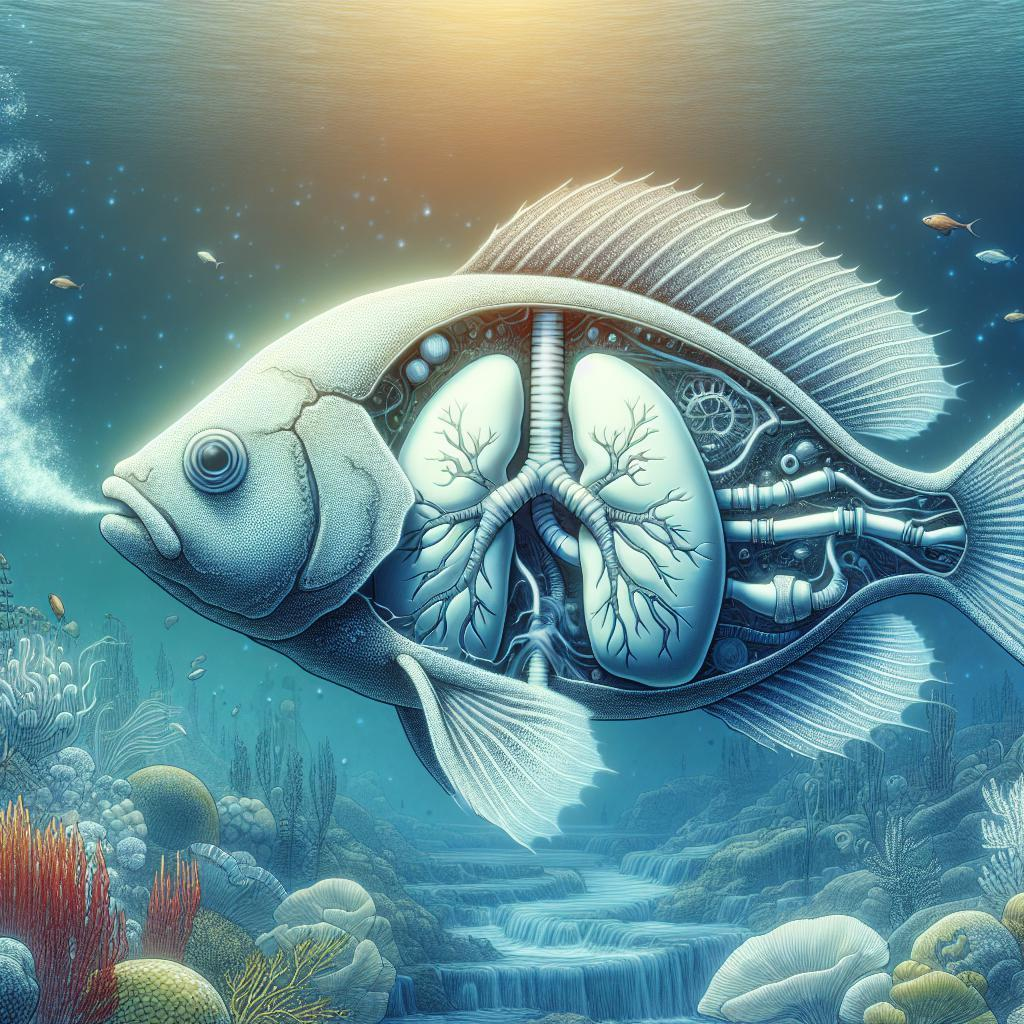}
{Please answer with one word: Gills allow fish to breathe underwater. Based on the image, what helps the fish to breathe underwater?}
{Lungs}
{
\textbf{<think>} The image shows a fish with its internal anatomy exposed, revealing a respiratory system that includes lungs and what appears to be gill-like structures. Given the problem's focus on what allows the fish to breathe underwater, and considering the visible structures within the fish, the respiratory system in the image is designed to mimic gills. While the image is stylized, it represents a form of adaptation to aquatic life, where the gill-like structures are the key to the fish's ability to breathe underwater. \textbf{</think>}\par
\boxed{\text{Gills}}
}
{
\textbf{<description>} The image shows a detailed cross-sectional view of a fish's body, revealing its internal organs and respiratory system. The lungs are prominently displayed, with visible branching structures resembling human lungs, and the fish is shown exhaling through its mouth, indicating the exchange of gases. The surrounding environment includes water, bubbles, and marine life, emphasizing the underwater setting. \textbf{</description>}\par
\textbf{<think>} The image depicts a fish with visible lungs and an open mouth, suggesting a method of gas exchange similar to what is seen in terrestrial lungs. While the question mentions gills, the lungs are clearly emphasized in the image. The fish appears to be breathing air through its lungs, which aligns with the presence of lungs in the cross-sectional view. The gills, while present in real fish, are not highlighted in this particular image. \textbf{</think>}\par
\boxed{\text{Lungs}}
}

\ViLPExample
{Tool use prior.}
{tab:tool-use-prior}
{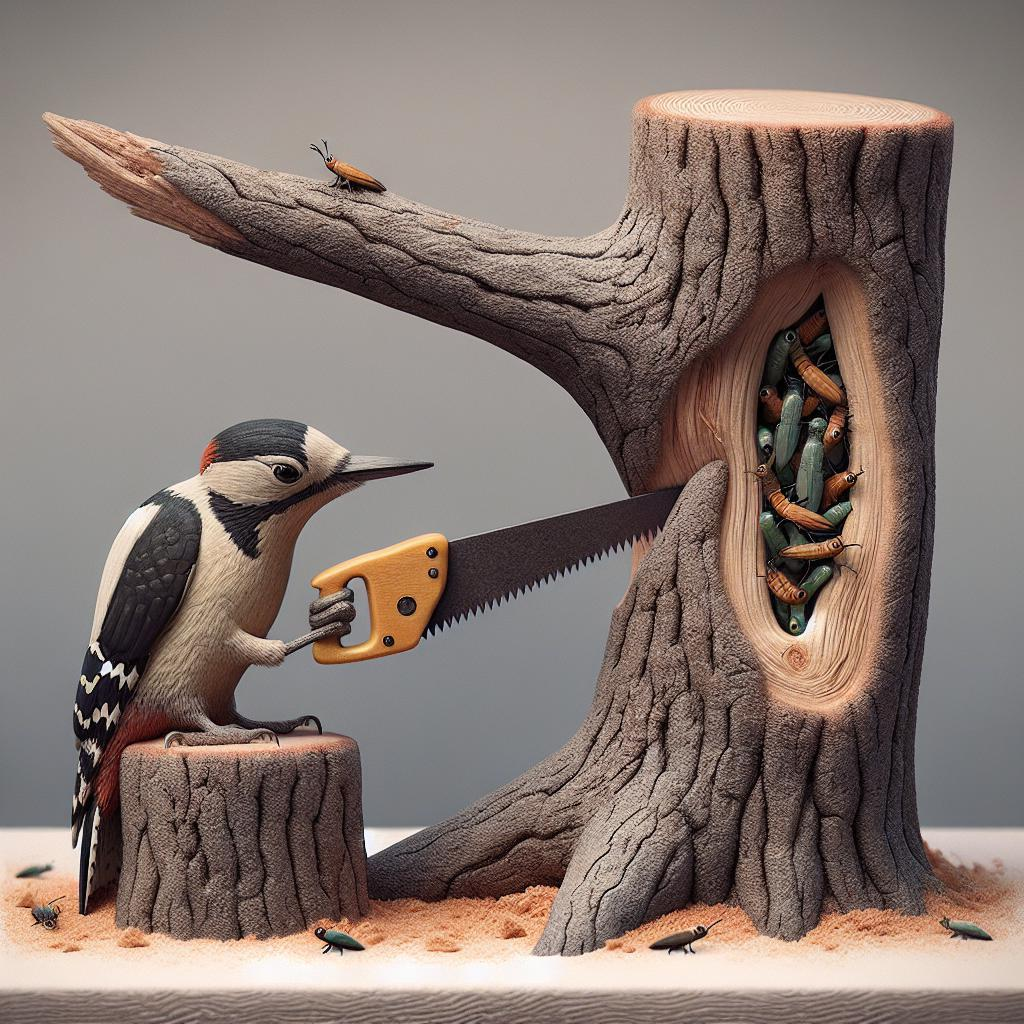}
{Please answer with one word: Woodpeckers usually use their beaks to search for insects in trees. According to the image, what do woodpeckers use to search for insects in trees?}
{Saw}
{
\textbf{<think>} The image shows a woodpecker using a hand saw to search for insects in a tree, which is an unusual and artificial depiction. Woodpeckers typically use their beaks for such tasks. \textbf{</think>}\par
\boxed{\text{Beaks}}
}
{
\textbf{<description>} The image shows a woodpecker using a hand saw to cut into a tree stump. The tree stump has a large hole revealing numerous insects inside. There are also a few insects on the tree branch above the stump and a few on the ground around the stump. The woodpecker appears to be searching for insects using the saw. \textbf{</description>}\par
\textbf{<think>} The image depicts a woodpecker using a hand saw to cut into a tree stump. This action contradicts the typical behavior of woodpeckers, which usually use their beaks to search for insects in trees. The presence of a saw implies that the woodpecker is not using its beak for the task at hand. The insects inside the tree stump and on the ground suggest that the woodpecker is indeed searching for them, but the tool (hand saw) indicates a departure from the usual method. \textbf{</think>}\par
\boxed{\text{Saw}}
}

\endgroup

Table~\ref{tab:vilp-qualitative-examples} shows five representative ViLP examples where the image conflicts with a common visual-language prior or with a fact stated in the question. In all cases, the correct answer should follow the image. These examples explain the ViLP results in the main text: \method improves Score because it is better at using the image when the image and the prior disagree, while it still keeps a high Prior score.

\paragraph{Landmark location.}
In Case~\ref{tab:landmark-location-prior}, the question reminds the model that the Statue of Liberty is in New York City. However, the image places the statue in a Paris scene, with the Eiffel Tower visible in the background. OPSD notices that the scene is unusual, but it treats the image as a manipulated scene and answers with the real-world prior, ``NYC''. \method first describes the visible Paris cues, especially the Eiffel Tower and the city background, and then answers ``Paris''. This case shows that \method uses the image as the main evidence when the question asks for the image-based location.

\paragraph{Text conflict.}
Case~\ref{tab:text-conflict} gives a simpler conflict. The question states the normal unit conversion, where one kilometer equals 1000 meters, but the image clearly writes ``1 km = 200 m''. OPSD reads this visual equation, but rejects it because it conflicts with the standard conversion, and finally answers ``1000''. \method follows the phrase ``as indicated in the image'' and answers ``200''. This example shows that the OPSD error is not only an OCR failure. OPSD can read the image text, but it does not give the image enough weight when the image conflicts with the prior.

\paragraph{Size prior.}
Case~\ref{tab:size-prior} tests a common animal-size prior. The question states that giraffes are very tall, but the image shows an elephant that appears taller than the giraffe in the current scene. OPSD follows the general prior that giraffes are tall and answers ``Giraffe''. \method compares the visible sizes of the two animals in the image and answers ``Elephant''. This case shows that \method is not only recognizing objects, but also using the visual relation between objects in the current image.

\paragraph{Biology prior.}
Case~\ref{tab:biology-prior} tests a biology prior. The question states that gills allow fish to breathe underwater, but the image highlights lung-like organs inside the fish. OPSD returns the common answer ``Gills''. \method describes the visible internal organs and answers ``Lungs''. The key point is that the question asks what helps the fish breathe underwater based on the image, so the visually highlighted organs should decide the answer.

\paragraph{Tool-use prior.}
Case~\ref{tab:tool-use-prior} tests a tool-use prior. Woodpeckers usually use their beaks to search for insects in trees, and the question states this prior. However, the image shows a woodpecker using a hand saw to open the tree trunk. OPSD answers ``Beaks'', which matches the prior but not the image. \method describes the saw and the insects in the tree, and answers ``Saw''. This is a clear case where the model must choose the current visual scene instead of normal world knowledge.

\paragraph{Summary.}
Across these examples, OPSD shows a consistent failure pattern. It often mentions the key visual evidence in its reasoning, but its final answer is still pulled back to the common prior or the fact stated in the question. This means the problem is not only perception. The model may see the relevant visual cue, but it does not always give that cue enough weight when making the final decision. This matches the shortcut risk discussed in the main text: answer-conditioned supervision over the whole trajectory can encourage answer-compatible reasoning that is not fully grounded in the image.

\method reduces this mismatch by separating the two steps. The model first writes a visual description, and this part is supervised by an image-only perception teacher. The privileged reasoning teacher is used only after this visual prefix is already in place. As a result, the later reasoning and final answer are more likely to use the image evidence rather than ignore it. These qualitative cases support the main claim of \method: the method does not simply remove useful priors. Instead, it helps the model decide when the image should override the prior.

\section{Additional Implementation Details}
\label{app:training_details}

This section reports the implementation details needed to reproduce both training and evaluation. We first list the prompts used by the student and the three teacher roles during training. We then give the training hyperparameters and the evaluation decoding configuration.

\subsection{Training Prompts}
\label{app:training_prompts}

For prompts that contain text, the image is placed before the text as an
image part in the same user message. Thus, the text box below is not sent
alone; it follows the image in the same user turn. The placeholders
\texttt{\{problem\}} and \texttt{\{reference\_solution\}} are replaced by
the current question and the privileged reference target. When a dataset
provides only a normalized answer, that answer is used as the target
string; when it provides a longer solution, the solution text is used.

% Visual prompt boxes used in Appendix training-prompt examples.
% They follow the red/green prompt-card style in the reference figure while
% keeping the prompt text verbatim through the listings engine.
\definecolor{PromptStudentFrame}{HTML}{A40000}
\definecolor{PromptStudentTitle}{HTML}{F4C7C7}
\definecolor{PromptStudentBack}{HTML}{F8DDDD}
\definecolor{PromptTeacherFrame}{HTML}{008A12}
\definecolor{PromptTeacherTitle}{HTML}{D8EBD7}
\definecolor{PromptTeacherBack}{HTML}{EAF5EA}
\definecolor{PromptShadow}{HTML}{B7B7B7}

\lstdefinestyle{promptlisting}{%
    basicstyle={\footnotesize\ttfamily},
    numbers=none,
    breaklines=true,
    breakatwhitespace=false,
    columns=fullflexible,
    keepspaces=true,
    showstringspaces=false,
}

\newtcblisting{studentpromptbox}{%
    enhanced,
    breakable,
    listing only,
    listing engine=listings,
    colback=PromptStudentBack,
    colframe=PromptStudentFrame,
    colbacktitle=PromptStudentTitle,
    coltitle=PromptStudentFrame,
    title={Student Prompt},
    fonttitle=\bfseries\Large,
    lefttitle=8pt,
    righttitle=8pt,
    toptitle=4pt,
    bottomtitle=4pt,
    left=8pt,
    right=8pt,
    top=7pt,
    bottom=7pt,
    boxrule=1.2pt,
    arc=2mm,
    outer arc=2mm,
    boxsep=0pt,
    before skip=6pt,
    after skip=10pt,
    drop shadow={PromptShadow},
    listing options={style=promptlisting}
}

\newtcblisting{teacherpromptbox}[1]{%
    enhanced,
    breakable,
    listing only,
    listing engine=listings,
    colback=PromptTeacherBack,
    colframe=PromptTeacherFrame,
    colbacktitle=PromptTeacherTitle,
    coltitle=PromptTeacherFrame,
    title={#1},
    fonttitle=\bfseries\Large,
    lefttitle=8pt,
    righttitle=8pt,
    toptitle=4pt,
    bottomtitle=4pt,
    left=8pt,
    right=8pt,
    top=7pt,
    bottom=7pt,
    boxrule=1.2pt,
    arc=2mm,
    outer arc=2mm,
    boxsep=0pt,
    before skip=6pt,
    after skip=10pt,
    drop shadow={PromptShadow},
    listing options={style=promptlisting}
}

\newtcolorbox{teacherpromptnote}[1]{%
    enhanced,
    breakable,
    colback=PromptTeacherBack,
    colframe=PromptTeacherFrame,
    colbacktitle=PromptTeacherTitle,
    coltitle=PromptTeacherFrame,
    title={#1},
    fonttitle=\bfseries\Large,
    lefttitle=8pt,
    righttitle=8pt,
    toptitle=4pt,
    bottomtitle=4pt,
    left=8pt,
    right=8pt,
    top=7pt,
    bottom=7pt,
    boxrule=1.2pt,
    arc=2mm,
    outer arc=2mm,
    boxsep=0pt,
    before skip=6pt,
    after skip=10pt,
    drop shadow={PromptShadow},
    fontupper=\footnotesize\ttfamily
}
\paragraph{Student rollout prompt.}
The student sees the image and the problem, but never sees the reference
target.

\begin{studentpromptbox}
Problem: {problem}

You are tasked with analyzing an image to generate a detailed description that can help you answer the question. First analyze the image and produce a self-contained description, detailed enough to lead to the correct answer. Do not include the final answer in the description. Wrap the entire description in <description> </description> tags.

Next, engage in an internal dialogue and include self-reflection or verification in your reasoning process. Provide detailed, step-by-step reasoning based on the image description and the image, and enclose this part within <think> </think> tags.

Finally, provide a single word or phrase answer to the question in \boxed{}.

The output format should be: <description> image description here </description> <think> reasoning process here </think> \boxed{FINAL ANSWER here}.
\end{studentpromptbox}

\paragraph{Image-only perception teacher prompt.}
The perception teacher is intentionally image-only in its external context. It does not receive the
problem text, answer options when present, or reference target. During token-level scoring, it still conditions on the student's already generated prefix $h_t$, as defined in the method section. Therefore, the perception loss on the description segment comes
from a teacher whose external evidence is only the image. This matches the role of $q_{\mathrm{img}}$ in the method section.

\begin{teacherpromptnote}{Image-only Perception Teacher Input}
<image only>

No extra problem text or reference target is provided outside the student prefix.
\end{teacherpromptnote}

\paragraph{Privileged reasoning teacher prompt.}
The reasoning teacher sees the image, the problem, and the reference
target. This teacher can use the reference target to guide the
reasoning path, but it does not supervise the description tokens.

\begin{teacherpromptbox}{Privileged Reasoning Teacher Prompt}
Problem: {problem}

Here is a reference solution to this problem:
=== Reference Solution Begin ===
{reference_solution}
=== Reference Solution End ===

After reading the reference solution above, make sure you understand the reasoning behind each step, and do not copy or paraphrase it. Now, using your own words and independent reasoning, derive the same final answer to the problem above. Think step by step, explore different approaches, and do not be afraid to backtrack or reconsider if something does not work out:

Please reason step by step, and put your final answer within \boxed{}. The output format should be: <think> reasoning process here </think> \boxed{FINAL ANSWER here}.
\end{teacherpromptbox}

\paragraph{Reference teacher prompt.}
The reference teacher uses the same privileged information as the
reasoning teacher, but it asks for the full structured output. In training,
we only use this teacher as the reference fallback loss signal for invalid
rollouts. This keeps the full privileged prompt from becoming the default
teacher for every token, while still giving a clear format recovery signal.

\begin{teacherpromptbox}{Reference Teacher Prompt}
Problem: {problem}

Here is a reference solution to this problem:
=== Reference Solution Begin ===
{reference_solution}
=== Reference Solution End ===

After reading the reference solution above, make sure you understand the reasoning behind each step, and do not copy or paraphrase it. Now, using your own words and independent reasoning, derive the same final answer to the problem above. Think step by step, explore different approaches, and do not be afraid to backtrack or reconsider if something does not work out:

Please first write a visual description that can lead to the correct answer. Wrap the entire description in <description> </description> tags. Then reason step by step, and enclose this part within <think> </think> tags. Finally, put your final answer within \boxed{}.

The output format should be: <description> image description here </description> <think> reasoning process here </think> \boxed{FINAL ANSWER here}.
\end{teacherpromptbox}

\subsection{Training Hyperparameters}
\label{app:training_hyperparams_subsec}

Table~\ref{tab:training_hyperparams} summarizes the main training
hyperparameters of \method. The two model scales use the same training
data, optimizer, learning rate, rollout sampling settings, loss weights,
and effective batch size.

\begin{table*}[ht]
\centering
\footnotesize
\setlength{\tabcolsep}{4.5pt}
\caption{Training hyperparameters of \method.}
\label{tab:training_hyperparams}
\begin{tabular}{lcc}
\toprule
Parameter & \method-3B & \method-7B \\
\midrule
Base model
& Qwen2.5-VL-3B-Instruct
& Qwen2.5-VL-7B-Instruct \\
Training epochs
& 1
& 1 \\
GPUs
& 8$\times$A100
& 8$\times$A100 \\
Effective batch size
& 32
& 32 \\
Optimizer
& Fused AdamW
& Fused AdamW \\
Learning rate
& $5\times10^{-6}$
& $5\times10^{-6}$ \\
LR scheduler
& Linear
& Linear \\
Maximum gradient norm
& 0.1
& 0.1 \\
Precision
& bf16
& bf16 \\
Distributed training
& ZeRO-2
& ZeRO-2 \\
Maximum prompt length
& 32,768
& 32,768 \\
Maximum completion length
& 4,096
& 4,096 \\
LoRA rank
& 64
& 64 \\
LoRA alpha
& 128
& 128 \\
LoRA dropout
& 0.05
& 0.05 \\
LoRA target modules
& \multicolumn{2}{c}{{q\_proj}, {k\_proj}, {v\_proj}, {o\_proj}, {gate\_proj}, {up\_proj}, {down\_proj}} \\
Rollout temperature
& 1.1
& 1.1 \\
Top-$p$ / Top-$k$
& 0.95 / 20
& 0.95 / 20 \\
$\lambda_{\mathrm{perc}}$
& 1.0
& 1.0 \\
$\lambda_{\mathrm{rea}}$
& 1.0
& 1.0 \\
$\lambda_{\mathrm{ref}}$
& 2.0
& 2.0 \\
Distillation temperature
& 1.0
& 1.0 \\
KL clipping
& 0.05
& 0.05 \\
\bottomrule
\end{tabular}%
\end{table*}

\subsection{Evaluation Configuration}

We also report the evaluation-time generation setting. This setting is separate from the training rollout setting in Table~A.III. The separation is important because \method changes the training supervision, but it does not use any extra information at test time. During evaluation, all teacher models, reference solutions, and segment masks are removed. The model receives only the image, the question, and the prompt used in the current evaluation setting.

The same decoding configuration is used for Baseline, OPSD, \method, and all ablation variants. We do not tune the decoding parameters for different models or benchmarks. This keeps the comparison focused on the learned model behavior, rather than on different test-time sampling choices. The same setting is also used in the same-prompt comparison in Appendix~E; that experiment changes the prompt control, not the decoding rule.

\begin{table}[t]
\centering
\footnotesize
\setlength{\tabcolsep}{4.5pt}
\caption{{Evaluation decoding configuration.} These settings are used for all reported evaluation results.}
\label{tab:evaluation_config}
\begin{tabular}{lc}
\toprule
Parameter & Value \\
\midrule
Maximum generated tokens & 4,096 \\
Number of samples per question & 5 \\
Temperature & 1.0 \\
Top-$p$ & 0.90 \\
Top-$k$ & 20 \\
Random seed & 42 \\
\bottomrule
\end{tabular}
\end{table}

For the eight main benchmarks, we generate five stochastic responses for each example using the decoding setting in Table~\ref{tab:evaluation_config}. Pass@5 is counted as correct if at least one of the five extracted answers is correct. Avg@5 is the mean correctness of the five responses. Thus, Pass@5 measures whether the model can find a correct answer within 5 trials, while Avg@5 measures the stability of the sampled answers. No reranking or manual selection is used.

For ViLP, we generate a single response per prompt using the same decoding settings. We then compute Score and Prior following the benchmark definition. Score measures performance on visually diagnostic questions, where the image may conflict with a common visual-language prior. Prior measures performance on prior-aligned questions, where the common prior is correct. This aligns with the goal of \method: the model should rely on the image when it matters, while still retaining useful prior-aligned knowledge.

For answer extraction, we use the official parser for each benchmark when available. For structured outputs that contain \texttt{\textbackslash boxed\{\}}, we use the content of the last box as the final answer. If an output has no parseable answer, it is counted as incorrect. We do not manually edit, complete, or correct model outputs before scoring. For \method, the 4,096-token limit applies to the entire generated sequence, including the visual description, reasoning process, and final answer.

\end{document}